%% file: submission.tex
\documentclass[10pt,twocolumn,letterpaper]{article}
\pdfoutput=1
\usepackage{iccv}
\usepackage{times}
\usepackage{epsfig}
\usepackage{graphicx}
\usepackage{amsmath}
\usepackage{amssymb}
\usepackage{subfigure}
\usepackage{tikz}
\usetikzlibrary{positioning}
\usetikzlibrary{intersections}

\newcommand\norm[1]{\left\lVert#1\right\rVert}
\newcommand{\edit}[1]{#1}

\usepackage{url}
\makeatletter
\g@addto@macro{\UrlBreaks}{\UrlOrds}
\makeatother
\iccvfinalcopy

\usepackage[pagebackref=true,breaklinks=true,letterpaper=true,colorlinks,bookmarks=false]{hyperref}

\def\iccvPaperID{0140} 

\ificcvfinal\fi
\begin{document}

\title{3D Time-Lapse Reconstruction from Internet Photos}

\author{Ricardo Martin-Brualla$^1$ \quad \quad \quad David Gallup$^2$ \quad \quad \quad  Steven M. Seitz$^{1,2}$\\
University of Washington$^1$ \quad \quad \quad Google Inc.$^2$\\
{\tt\small \{rmartin,seitz\}@cs.washington.edu} \quad \quad {\tt\small dgallup@google.com} 
}

\maketitle

\input{abstract}

\input{intro}

\input{related_work}
\input{overview}

\input{depthmap}

\input{appearance}

\input{implementation}
\input{results}

\input{conclusion}
\clearpage

{\small
\bibliographystyle{ieee}

\input{submission.bbl}
}

\end{document}

%% file: abstract.tex
\begin{abstract}
Given an Internet photo collection of a landmark,
we compute a 3D time-lapse video sequence where a
virtual camera moves continuously in time and space.
While previous work assumed a static camera, the
addition of camera motion during the time-lapse creates
a very compelling impression of parallax.  Achieving
this goal, however, requires addressing multiple technical
challenges, including solving for time-varying depth maps, 
regularizing 3D point color profiles over time, and reconstructing
high quality, hole-free images at every frame from the 
projected profiles.  
Our results show photorealistic time-lapses of skylines and
natural scenes over many years, with dramatic parallax effects.

\end{abstract}

%% file: intro.tex
\section{Introduction}

Time-lapses make it possible to see events that are otherwise 
impossible to observe, like the motion of stars in the night sky 
or the rolling of clouds.
By placing fixed cameras, events over even longer time spans can be imaged, like
the construction of skyscrapers or the retreat of glaciers~\cite{ExtremeIceSurvey}.
Recent work~\cite{TimelapseMining,MatzenECCV14} has shown the
exciting possibility of computing time-lapses from large Internet photo collections.
In this work, we seek to compute 3D time-lapse video sequences from Internet photos where a
virtual camera moves continuously in both time and space.

Professional photographers exploit small camera motions to
capture more engaging time-lapse sequences~\cite{LaforetTimelapseTutorial}.
By placing the camera on a controlled slider platform
the captured sequences show compelling parallax effects.
Our technique allows us to recreate such cinematographic effects
by simulating virtual camera paths, but with Internet photo collections.

\begin{figure}

\begin{tikzpicture}
\usetikzlibrary{shapes,snakes}

    \begin{scope}
        \node[anchor=south west,inner sep=0] (internetimages) at (0.0,0) {\includegraphics[height=28mm]{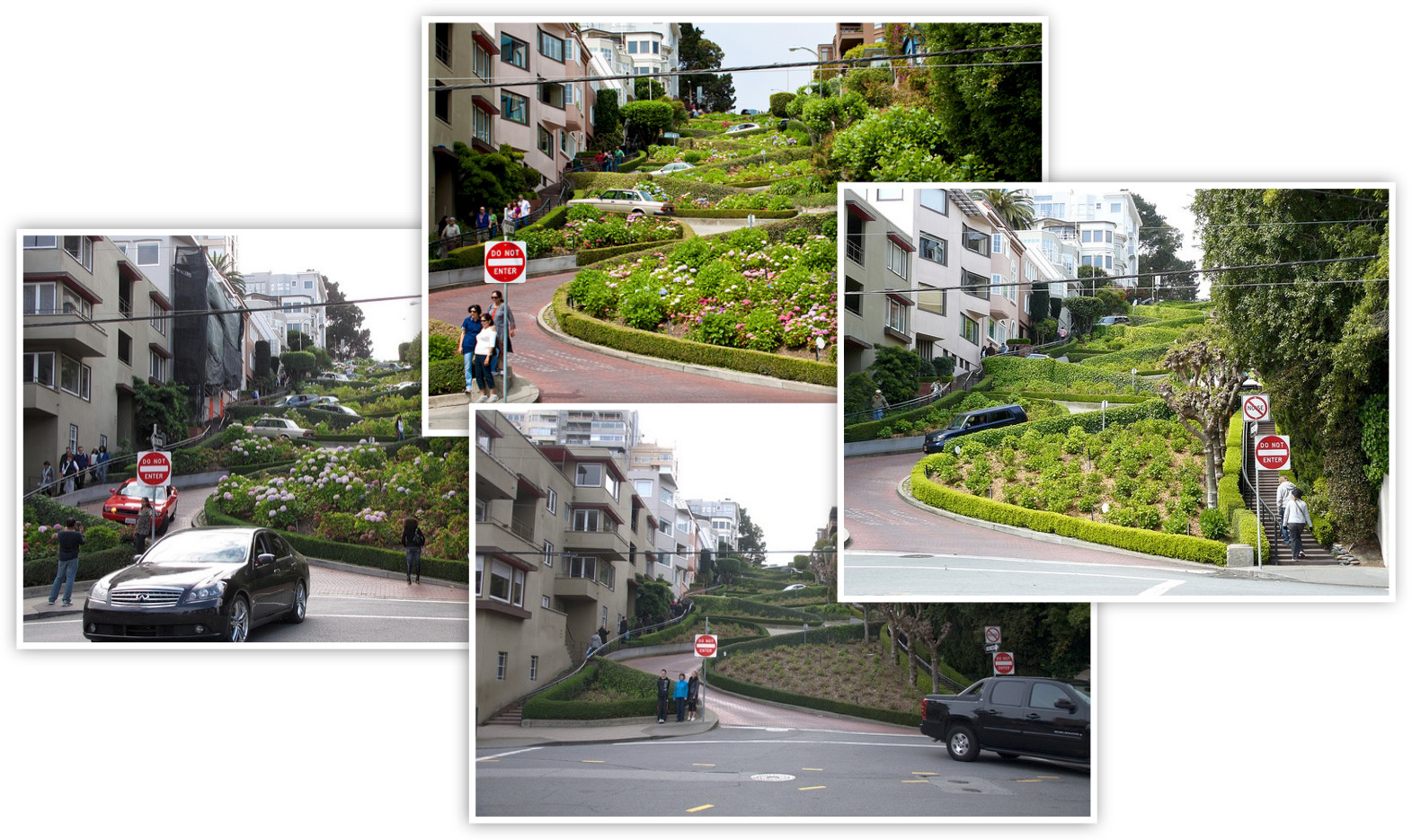}};
        \node[anchor=south west,inner sep=0] (internetphotostitle) at (1.65	, -0.35) {\small Internet Photos};
        
        \node[anchor=south west,inner sep=0] (diagram) at (4.9,0.15) {\includegraphics[height=25mm]{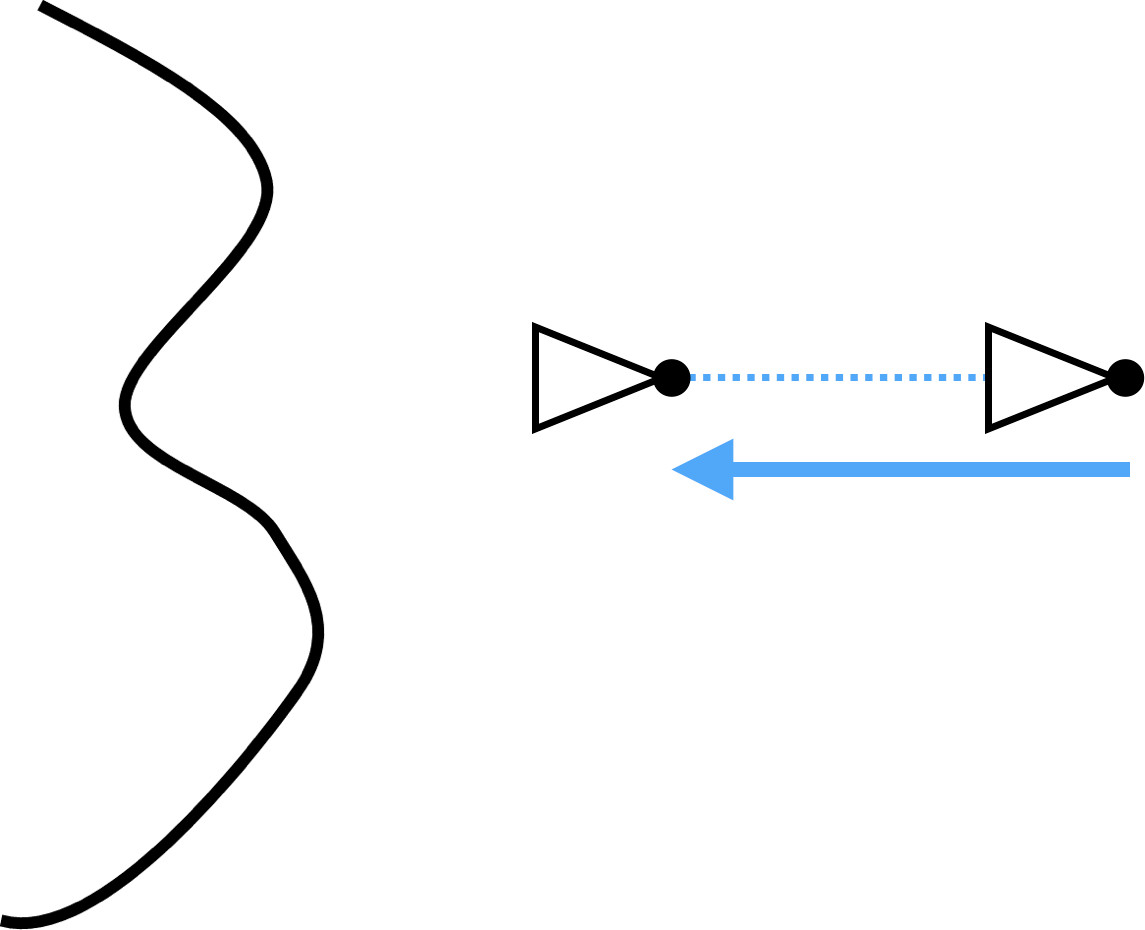}};
        \node[anchor=south west,inner sep=0] (medium_cap) at(4.85,-0.35) {\small 3D Scene};
        \node[anchor=south west,inner sep=0, text width=1.8cm, align=center] (medium_cap) at(6.4, 0.5) {\small Virtual \mbox{Camera} Path};

        \node[anchor=north west,inner sep=0] (filmstrip) at(0, -0.7) {\includegraphics[width=1\linewidth]{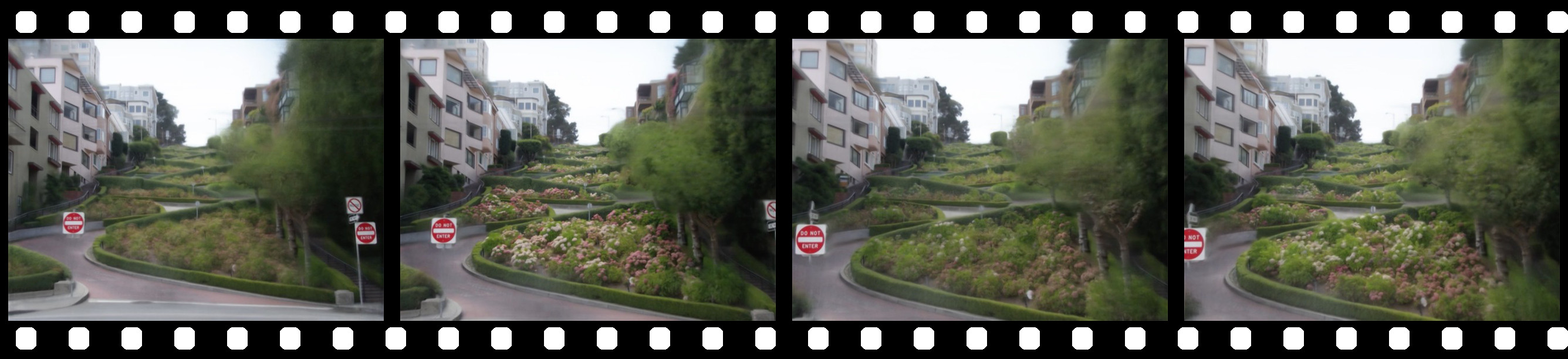}};
        \node[below of=filmstrip,anchor=north] (medium_cap)  {\small Synthesized 3D Time-lapse};
    \end{scope}
\end{tikzpicture}

\caption{
In this paper we introduce a technique to produce high quality
3D time-lapse movies from Internet photos, where a virtual
camera moves continuously in space during a time span of several years.
Top-left: Sample input photos of the gardens in Lombard Street, San Francisco.
Top-right: Schematic of the 3D scene and the virtual camera path.
Bottom: Example frames of the synthesized 3D time-lapse video.
Please see the supplementary video available at the project website~\cite{3DTimelapseWebsite}. 
Credits: Creative Commons photos from Flickr users Eric Astrauskas, Francisco Antunes, Florian Plag and Dan Dickinson.
}
\label{fig:teaser-figure}
\end{figure}
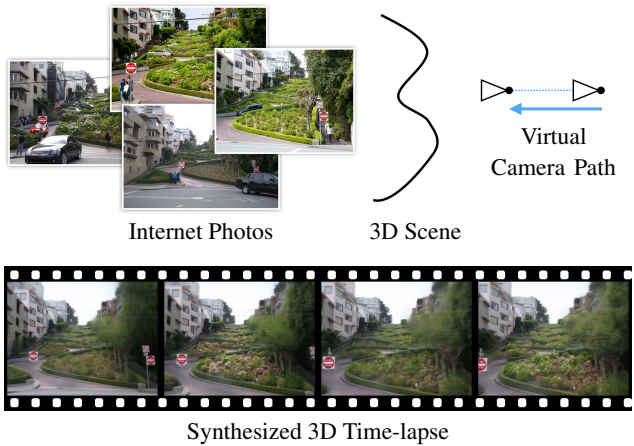

We build on our previous work~\cite{TimelapseMining} and introduce
key new generalizations that account for time-varying geometry and
enable virtual camera motions. 
Given a user-defined camera path through space and over time, 
we first compute time-varying depthmaps
for the frames of the output sequence.
Using the depthmaps, we compute correspondences across the image
sequence (aka. ``3D tracks'').
We then regularize the appearance of each track over time (its ``color profile'').
Finally, we reconstruct the time-lapse video frames from the projected
color profiles.

Our technique works for any landmark that is widely photographed,
where, over time, thousands of people have taken photographs of roughly the same view.
Previous work~\cite{TimelapseMining} identified more than 10,000
such landmarks around the world.

The key contributions of this paper are the following:
1) recovering time-varying, temporally consistent depthmaps from Internet photos via a more robust 
adaption of~\cite{VideoDepthmapRecovery2009}, 2)
a 3D time-lapse reconstruction method that solves for the temporal
color profiles of 3D tracks, and 3)
an image reconstruction method that computes hole-free output frames from
projected 3D color profiles.
Together, these contributions allow our system to correctly handle changes in geometry and camera position, yielding time-lapse results superior to those of~\cite{TimelapseMining}.

%% file: related_work.tex
\section{Related Work}

\edit{Our} recent work~\cite{TimelapseMining} introduced a method to
synthesize time-lapse videos from Internet Photos
spanning several years.
The approach assumes a static scene and recovers
one depthmap that is used to warp the input images into
a static virtual camera. 
A temporal regularization over individual pixels of the output
volume recovers a smooth appearance for the whole sequence.
The static scene assumption proved to be a failure mode of that
approach resulting in blurring artifacts when scene geometry changes.
We address this problem by solving for time-varying geometry,
and extend
the appearance regularization to 3D tracks and moving camera paths.

Very related to our work, Matzen and Snavely~\cite{MatzenECCV14} 
model the appearance of a scene over time
from Internet photos
by discovering space-time cuboids, corresponding to rectangular surfaces in the scene
visible for a limited amount of time, like
billboards or graffiti art.
Similarly, the 4D Cities project~\cite{SchindlerCVPR2010, SchindlerCVPR2007} models
the changes in a city over several decades using historical imagery.
By tracking the visibility of 3D features over time, they are able to reason
about missing and inaccurate timestamps.
In contrast, we synthesize photorealistic
time-lapses of the scene, instead of sparse 4D representations
composed of textured rectangular patches or 3D points.

Photobios~\cite{Photobios} are 
visualizations computed from personal photo
collections that show how people age through time. 
The photos are displayed one by one, while fixing the location
of the subject's face over the whole sequence.
These visualizations are limited to faces and do not create the
illusion of time flowing continuously, like our time-lapse sequences do.

Parallax Photography, by Zheng \etal~\cite{ParallaxPhotography},
creates content-aware camera paths that optimize for
parallax effects in carefully collected datasets.
Additionally, Snavely \etal~\cite{FindingPaths} discover orbit paths that
are used to navigate Internet photo collections more efficiently.
In our work, the user specifies the camera path as input.

Modeling the appearance of a scene from Internet photos is challenging,
as the images are taken with different illumination, at different times of
day and present many occluders.
Laffont \etal~\cite{CoherentIntrinsicImages} regularize the appearance
of a photo collection by computing coherent intrinsic images across the collection.
Shan \etal~\cite{VisualTuringTest} detect cloudy images in a photo collection, 
to initialize a factored lighting model for a 3D model recovered from Internet photos.

Generating time-lapse videos from static webcams has also been studied in prior work.
Bennett and McMillan~\cite{ComputationalTimelapse} propose several objective functions
to synthesize time-lapse videos, that showcase different aspects of the changes
in the scene.
Rubinstein \etal \cite{MotionDenoising} reduce flicker caused by small motions in time-lapse sequences.

Kopf \etal~\cite{FirstPersonHyperlapse} generate smooth hyper-lapse videos from first-person footage.
Their technique recovers scene geometry to stabilize the video sequence, synthesizing views
along a smoothed virtual camera path that allows for faster playback.

Although multi-view stereo has been an active topic of research for
many years~\cite{MultiViewStereoSeitz06}, few works have looked into time-varying 
reconstruction outside of carefully calibrated datasets.
Zhang \etal~\cite{ZhangSpaceTimeStereo} reconstruct time-varying depthmaps
of moving objects with a spacetime matching term. 
Larsen \etal~\cite{Larsen2007} compute temporally consistent depthmaps
given calibrated cameras using optical flow to enforce
depth consistency across frames.
Zhang \etal~\cite{VideoDepthmapRecovery2009} introduce a method to recover
depthmaps of a static scene from handheld captured video sequences.
Their method first computes a 3D pose for every frame, and then
jointly optimizes the depthmaps for every frame,
using a temporal consistency term. 
We extend their approach to
 handle dynamic scenes, and adapting it to Internet photo collections.







%% file: overview.tex
\section{Overview}

\begin{figure*}[ht!]
\centering
\includegraphics[width=0.32\linewidth]{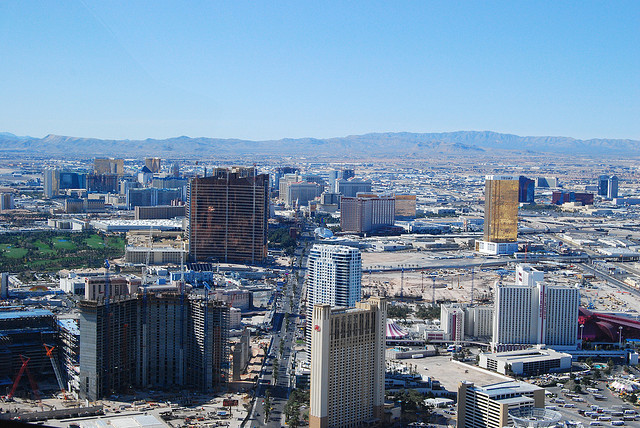}
\includegraphics[width=0.32\linewidth]{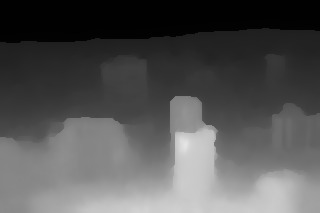}
\includegraphics[width=0.32\linewidth]{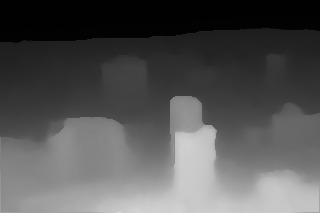}\\
\vspace{-4pt}%
\subfigure[Sample input photos]{%
\includegraphics[width=0.32\linewidth]{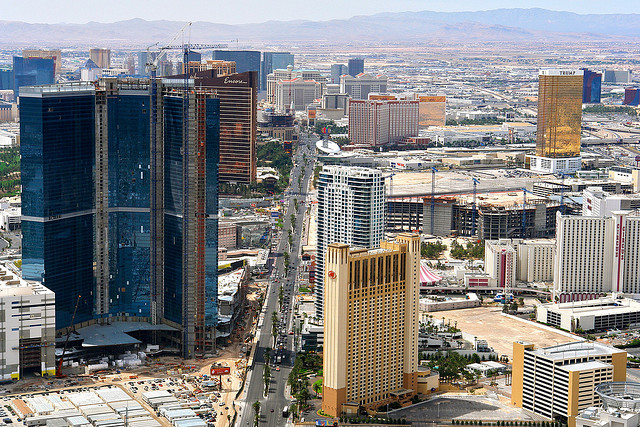}}
\subfigure[Initialized depthmap]{%
\includegraphics[width=0.32\linewidth]{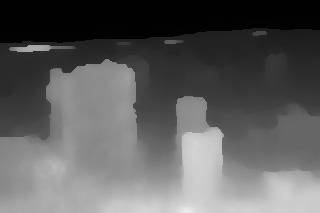}}
\subfigure[After joint optimization]{%
\includegraphics[width=0.32\linewidth]{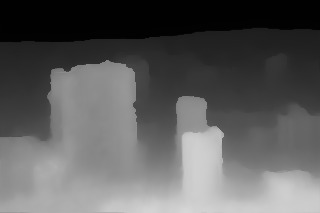}}%
\caption{
Results of our time-varying depthmap reconstruction.
a) Sample input photos at different times from the Las Vegas skyline scene (not aligned to virtual camera).
b) Initialized depthmap for the corresponding time of the photos on the left.
c) Jointly optimized depthmaps.
Note that artifacts near the top in the second depthmap are fixed after the joint optimization.
The improvements to temporal consistency are dramatic and better seen in the supplementary video~\cite{3DTimelapseWebsite}.
Credits: Creative Commons photos from Flickr users Butterbean and Alex Proimos.
}
\label{fig:depthmap_opt}
\end{figure*}

Given an Internet photo collection of a landmark, we 
seek to compute time-lapse video sequences where a
virtual camera moves continuously in time and space. 
As a preprocessing step, we compute the 3D pose
of the input photo collection using Structure-from-Motion (SfM)
techniques~\cite{agarwal2011romeinaday}.

First, a user specifies a desired virtual camera path through the reconstructed scene.
This can be defined by specifying a reference camera and a parameterized
motion path, such as an orbit around a 3D point or a ``push'' or ``pull'' motion path~\cite{LaforetTimelapseTutorial}.
Good reference cameras are obtained using the scene
summarization approach of~\cite{SceneSummarization}.

Our system starts by computing time-varying, temporally consistent depthmaps for all
output frames in the sequence, as described in Section~\ref{sec:depthmap}.
Section~\ref{sec:appearance} introduces our novel 3D time-lapse reconstruction,
that computes time-varying, regularized color profiles for 3D tracks
in the scene.
We then present a method to reconstruct output video frames from the projected color profiles.
Finally, implementation details are described in Section~\ref{sec:implementation}
and results are shown in Section~\ref{sec:results}.

For the rest of the paper, we only consider
images whose cameras in the 3D reconstruction are close
to the reference camera.
We use the same criteria for image
selection as~\cite{TimelapseMining},
that selects cameras by comparing their optical axis and camera center 
to those of the reference camera.

Throughout this paper, we will use the following terminology:
each photo in the input collection  consists of an image $I_i$,
a registered camera $C_i$ and a timestamp $t_i$.
We also define the sequence $\mathcal{I} = (I_1,\ldots,I_N)$ as the chronologically
sorted input image sequence.
The output 3D time-lapse sequence is composed of $M$ output frames whose views $V_j$
are equally spaced along the virtual camera path
and span the temporal extent of the input sequence, from earliest to the latest photo.

%% file: depthmap.tex

\section{Time-Varying Depthmap Computation}
\label{sec:depthmap}

In this section we describe how to compute a temporally consistent depthmap
for every view in the output sequence.
The world  changes in different ways over time spans 
of years compared to time spans of seconds.
In multi-year time scales, geometry changes by adding or substracting surfaces, like
buildings being constructed or plants growing taller,
and we design our algorithm to account for such changes.

Recovering geometry from Internet photos is challenging, as these photos are captured with different cameras,
different lighting conditions, and with many occluders.
A further complication is that included timestamps are often wrong, as noted in previous work~\cite{hauagge_bmvc2014_outdoor, MatzenECCV14}.
Finally, most interesting scenes undergo
changes in both texture and geometry, further complicating depthmap reconstruction.

\subsection{Problem Formulation}

Our depth estimation formulation is similar to that of~\cite{VideoDepthmapRecovery2009}, except
that we~1) use a Huber norm for the temporal consistency term to make it 
robust to abrupt changes in geometry, and~2) replace the photo-consistency term
with that of~\cite{TimelapseMining} which is also robust to temporally varying geometry
and appearance changes which abound in Internet photo collections.

We pose the problem as solving for a depthmap $D_j$ for each synthesized view $V_j$,
by minimizing the following energy function:
\begin{equation}
\label{eq:depthmap-formulation}
\sum_j{\left[E^{d}(D_j)+\alpha E^{s}(D_j)\right]} + \sum_{j, j'}{\beta_{j,j'}E^{t}(D_j, D_{j'})}
\end{equation}
where $E^{d}$ is a data term based on a matching cost volume, $E^{s}$ is a spatial regularization term between neighboring pixels,
and $E^{t}$ is a 
binary temporal consistency term that enforces the projection
of a neighboring depthmap $D_{j'}$ into the view $V_j$ to be consistent with $D_j$.
The binary weight $\beta_{j,j'}$ is non-zero only for close values of $j$ and $j'$.

Given the projected depthmap $D_{j'\rightarrow j}$ of the depthmap $D_{j'}$ into view $V_j$,
we define the temporal regularization term  for a pixel $p$ in $V_j$ as:
\begin{equation}
\label{eq:temporal_term}
E^t(D_j,D_{j'})(p) = \delta\left(D_j(p) -  D_{j'\rightarrow j}(p)\right)
\end{equation}

if there is a valid projection of $D_{j'}$ in view $V_j$ at $p$ and $0$ otherwise,
and where $\delta$ is a regularization loss.
We use z-buffering to project the depthmap so that the constraint is enforced only
on the visible pixels of view $V_j$. 
Zhang \etal~\cite{VideoDepthmapRecovery2009} assume a Gaussian prior 
on the depth of the rendered depthmap, equivalent to $\delta$ being the $L_2$ norm.
In contrast, our scenes are not static and present abrupt changes in depth,
that we account for by using a robust loss, the Huber norm.


The data term $E^d(D_j)$ is defined
as the matching cost computed from a subset of
input photos $\mathcal{S}_j \subset \mathcal{I}$ for each view $V_j$.
We choose the subset as the subsequence of length $l = 15\% \cdot N$
centered at the corresponding view timestamp.

Using the images in subset $\mathcal{S}_j$, we compute aggregate matching costs following~\cite{TimelapseMining}.
First, we generate a set of fronto-parallel
planes to the view $V_j$ using the computed 3D SfM reconstruction.
We set the range to cover all but the $0.5\%$ nearest
and farthest SfM 3D points from the camera.
In scenes with little parallax this approach might still fail, so we further discard SfM points that
have a triangulation angle of less than 2 degrees.

For each pixel $p$ in view $V_j$ and depth $d$, we compute
the pairwise matching cost $C^j_{a, b}(p, d)$ for images $I_a,I_b \in \mathcal{S}_j$,
by projecting both images onto the fronto-parallel plane at depth $d$
and computing normalized cross correlation with filter size $3 \times 3$.
We adapt the best-k strategy described in~\cite{Kang04extractingviewdependent}
to the pairwise matchings costs and define the aggregated cost as:
\begin{equation}
C^j(p, d) = \textbf{median}_{a\in \mathcal{S}_j}\left(\textbf{median}_{b\in \mathcal{S}_j}C^j_{a, b}(p, d) \right)
\end{equation}

Finally, the spatial regularization $E^s$ consists of the differences of
depth between 4 pixel neighborhoods, using the Huber norm,
with a small scale parameter to avoid the stair-casing effects observed by ~\cite{DTAM}.

\subsection{Optimization}
The problem formulation of Equation~\ref{eq:depthmap-formulation} is hard to solve directly, as the 
binary temporal regularization term ties the depth of pixels
across epipolar lines.
We optimize this formulation similarly to Zhang \etal~\cite{VideoDepthmapRecovery2009},
by first computing each depthmap $D_j$ independently, \ie, without the 
consistency term $E^t$, and then
performing coordinate descent, where the depthmap $D_j$
is optimized while the others are held constant.
We iterate the coordinate descent through all depthmaps for two
iterations, as the solution converges quickly.

\edit{We solve the problem} in the continuous domain with non-linear optimization~\cite{ceres-solver}, adapting
the data term to the continuous case by interpolating the cost values for a pixel at different
depths using cubic splines.
We initialize each individual depthmap $D_j$ by solving the MRF formulation of~\cite{TimelapseMining}
for its corresponding support image set $\mathcal{S}_j$.

The joint optimization produces more stable depthmaps that
exhibit fewer artifacts than the initialized ones without the temporal
consistency term.
Figure~\ref{fig:depthmap_opt} shows examples of recovered time-varying depthmaps.
The improvements in temporal consistency for the jointly optimized sequence are best seen in
the supplementary video~\cite{3DTimelapseWebsite}.

%% file: appearance.tex
\section{3D Time-Lapse Reconstruction}
\label{sec:appearance}

Our goal is to produce photorealistic time-lapse videos that visualize
the changes in the scene while moving along a virtual camera path.
We pose the 3D time-lapse reconstruction problem as recovering time-varying,
regularized color profiles for 3D tracks in the scene.
A 3D track is a generalization of an image-to-image feature correspondence, which accounts for changes in 3D scene structure, and occlusions between views (See Fig.~\ref{fig:motion-track}).
First, we generate 3D tracks by following correspondences induced by
the depthmap and the camera motion.
We then solve for the temporal appearance of each 3D track, by projecting
them onto the corresponding input images and solving for time-varying, regularized
color profiles.
Finally, we reconstruct the output time-lapse video from the projected color profiles
of the 3D tracks.

\subsection{Generating 3D Tracks}
\label{sec:motion-tracks}
\input{figure_motion_track_v3}

We generate 3D tracks that follow the flow induced in the output sequence by the time-varying
depthmap and the camera motion.
Ideally, a track represents a single 3D point in the scene, whose appearance
we want to estimate.
However, occlusions and geometry changes may cause a track
to cover multiple 3D points.
Since the appearance regularization described in the next subsection is
robust to abrupt changes in appearance, 
our approach works well even with occlusions.



A 3D track is defined by a sequence of 3D points $t=(q_{j_1},\ldots,q_{j_n})$ for corresponding output views  $j_1,\ldots,j_n$.
To generate a 3D track, we define first a 3D point $q$ for a view $V$ that lies
on the corresponding depthmap $D$.
Let $p'$ be the projection of the 3D point $q$ onto the next view $V'$.
We then define the track's next 3D point $q'$ as the backprojection of pixel $p'$ 
onto the corresponding depthmap $D'$.
We compute the next 3D point $q''$ by repeating this process from $q'$.
We define a whole track by iterating forwards and backwards in the sequence,
and we stop the track if the projection falls outside the current view.
3D tracks are generated so that the output views are covered with sufficient density as described in Section \ref{reconstructing-video}.

Figure~\ref{fig:motion-track} shows the 3D track generation process.
Note that when the geometry is static, points in a 3D track remain constant
thanks to the robust norm used in the temporal consistency term, that promotes
depthmap projections to match between frames.
While drift can occur through this chaining process, in practice
this does not affect the quality of the final visualizations.

\subsection{Regularizing Color Profiles}
\input{figure_subpixel_samples}

We want to recover a time-varying, regularized color profile for
each 3D track $t$.
This is challenging as Internet photos display a lot of variation in appearance
and often contain outliers, as noted in Section~\ref{sec:depthmap}.
We make the observation that the albedo of most surfaces in the real world
does not change rapidly, and its variability in appearance stems mostly
from illumination effects.
Intuitively, we would like our time-lapse sequences to reveal the infrequent texture
changes (the signal) while hiding the variability and outliers of the input photo collection (the noise).

To solve for time-varying color profiles,~\cite{TimelapseMining}
used a temporal regularization term with a robust norm, that
recovers piecewise continuous appearances of pixels in an output image sequence.
The approach is restricted to a static virtual camera, 
as it works on the 2D domain by regularizing each pixel in the output sequence
independently.
Our approach uses the same temporal term to regularize the color profile of each 3D track.

Given a 3D track $t = (q_{j_1},\ldots,q_{j_n})$,
we define its appearance in view $V_j$ as the RGB value $y^t_j \in [0,1]^3$.
To compute $y^t_j$, we first assign input images
to their closest view in time and denote these images assigned to view $V_j$ by the support set $\mathcal{S}_j'\subset\mathcal{I}$.
Note that the sets $\mathcal{S}_j'$ are not overlapping, whereas the support sets 
$\mathcal{S}_j$ used for depthmap computation are.
We then project the 3D point $q_j$ to camera $C_i$ using a z-buffer
with the depthmap $D_j$ to check for occlusions and 
define $x^t_i$ as the RGB value of image $i$ at the projection of $q_j$.

We obtain a time-varying, regularized color profile for each 3D track $t$ by minimizing the following energy
function:
\begin{equation}
\label{eq:temporal-reg}
\sum_j{\sum_{i \in \mathcal{S}_j'}}{\delta_{d}\left(\norm{x^t_i - y^t_j}\right)}
+ \lambda\sum_{j}{\delta_{t}\left(\norm{y^t_{j+1} - y^t_j}\right)}
\end{equation}
where the weight $\lambda$ controls the amount of regularization,
and both $\delta_d$ and $\delta_t$ are the Huber norm, 
to reduce the effects of outliers in $x^t_j$ and
promote sparse temporal changes in the color profile.

In contrast to~\cite{TimelapseMining}, the color profiles of the 3D tracks
do not correspond to pixels in the output frames.
We thus save the color profile $y^t$, together with the 2D projections $p^t_j$
of the track 3D points $q^t_j$ into the view $j$, as \emph{projected profiles}
that are used to reconstruct the output frames.
Figure~\ref{fig:subpixel-samples} shows a diagram of a projected color profile.

\input{figure_subpixel_recon}
\subsection{Reconstructing Video from Projected Profiles} \label{reconstructing-video}

Given regularized projected color profiles computed for a set of 3D tracks $\mathcal{T}$,
we seek to reconstruct output frames of the time-lapse video that best fit the recovered color profiles.

We cast the problem of reconstructing each individual frame
as solving for the image that best
matches the color values of the projected color profiles
when applying bilinear interpolation at the profiles' 2D projections.
Figure~\ref{fig:subpixel-recon} visualizes the reconstruction process,
where the output pixels' color values are related to the projected profiles' samples
by bilinear interpolation weights.

For a given output view $V_j$,
let $Y_{u,v}\in[0,1]^3$ be the RGB value of the pixel $(u,v) \in \mathbb{N}^2$ in the synthesized output frame $Y$.
Let the regularized projected profile  for a track $t$ at view $V_j$ have an RGB value $y^t$ and a 2D projection $p^t\in\mathbb{R}^2$.
We solve for the image $Y$ that minimizes
\begin{equation}
\sum_{t\in\mathcal{T}}{\norm{y^t - {\textstyle\sum_{s=1}^4 {w^t_s Y_{u^t_s,v^t_s}}}}_2}
\end{equation}
where $u^t_s,v^t_s$ are the integer coordinates of the 4 neighboring pixels to $p^t$
and $w^t_s$ their corresponding bilinear interpolation weights.

The reconstruction problem requires the set of 3D tracks $\mathcal{T}$
to be dense enough that every pixel $Y_{u,v}$ has a non-zero weight in the optimization,
\ie, each pixel center is within $1$ pixel distance of a projected profile sample.
To ensure this, we generate 3D tracks using the following heuristic:
we compute 3D tracks for all pixels $p$ in the middle view $j$ of the sequence,
so that the 3D track point $q^t_j$ projects to the center of pixel $p$ in $V_j$.
Then, we do the same for all pixels in the first and last frame.
Finally, we iterate through all pixels in the output frames $Y$
and generate new 3D tracks if there is no sample within
$\epsilon \leq 1$ pixels from the pixel center coordinates.

\setlength{\textfloatsep}{10pt}

\begin{figure}
\begin{tikzpicture}
    \node[anchor=south west,inner sep=0] (image) at (0,0) {\includegraphics[width=0.49\linewidth]{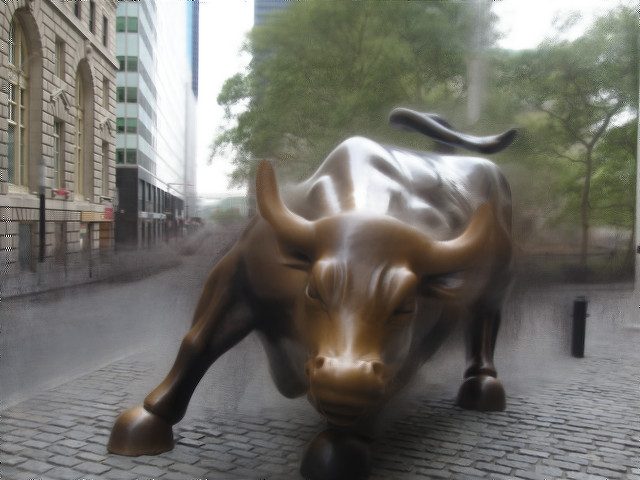}};
    \begin{scope}[x={(image.south east)},y={(image.north west)}]
        \node[anchor=south west,inner sep=0] (image) at (0.49,0.49) {\includegraphics[width=0.245\linewidth]{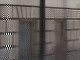}};
        \draw [semithick] (0.003,1-0.420833) rectangle (0.117188,1-0.535417);
        \draw [semithick] (0.49,0.49) rectangle (0.99,0.99);
        \draw [semithick,dashed] (0.003,1-0.420833) -- (0.49,0.99);
        \draw [semithick,dashed] (0.117188,1-0.535417) -- (0.99,0.49);
    \end{scope}
\end{tikzpicture}
\begin{tikzpicture}
    \node[anchor=south west,inner sep=0] (image) at (0,0) {\includegraphics[width=0.49\linewidth]{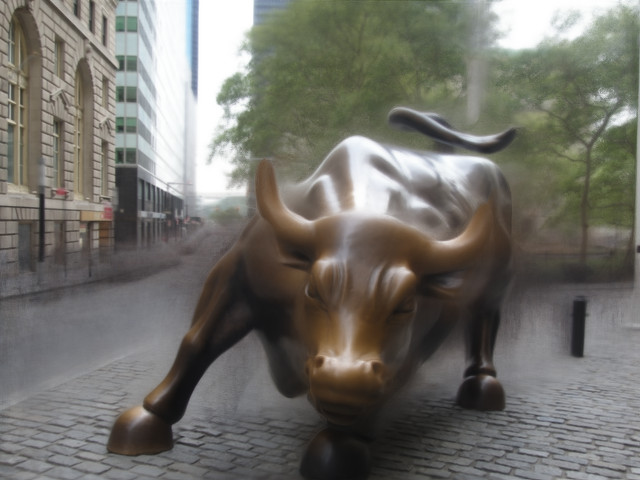}};
    \begin{scope}[x={(image.south east)},y={(image.north west)}]
        \node[anchor=south west,inner sep=0] (image) at (0.49,0.49) {\includegraphics[width=0.245\linewidth]{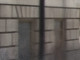}};
        \draw [semithick](0.003,1-0.420833) rectangle (0.117188,1-0.535417);
        \draw [semithick](0.49,0.49) rectangle (0.99,0.99);
        \draw [semithick,dashed] (0.003,1-0.420833) -- (0.49,0.99);
        \draw [semithick,dashed] (0.117188,1-0.535417) -- (0.99,0.49);
    \end{scope}
\end{tikzpicture}
\caption{
Comparison of different values of the 3D track sampling threshold $\epsilon$ for the Wall Street Bull scene.
Left: Artifacts are visible when $\epsilon=1$ pixel, with alternating black and white pixels,
 as the reconstruction problem is badly conditioned.
Right: Using $\epsilon=0.4$ pixel, the artifacts are not present.
}
\label{fig:subpixel-artifacts}
\end{figure}

The reconstruction problem can be badly conditioned, producing artifacts in
the reconstructions, such as contiguous pixels with alternating black and white colors.
This happens in the border regions of the image that have lower sample density.
We avoid such artifacts by using a low threshold value $\epsilon=0.4$ pixels, so
that for each pixel there is a projected profile whose bilinear interpolation weight is $>0.5$.
Figure~\ref{fig:subpixel-artifacts} shows an example
frame reconstruction using two different threshold values for $\epsilon$.

%% file: figure_motion_track_v3.tex
\begin{figure}
\centering

\subfigure[] {
\begin{tikzpicture}[scale=1,on grid]
\newcommand{\basestation}[4]{%
    \node[inner sep=0pt, circle, opacity=0, #3, minimum size=2pt, fill=red] (#1) at (0*120+#2:0.2cm) {};
    \path[draw, thick, #3] 
            (0*120+#2:0.2cm) --
            (1*120+#2:0.2cm) --
            (2*120+#2:0.2cm) -- cycle;
    \node[anchor=north,inner sep=2pt] at (#1) {{\footnotesize #4}};
}   

\basestation{C1}{255}{shift={(-.8, .1)}}{$V$};
\basestation{C2}{270}{shift={(0.0, 0)}}{$V'$};
\basestation{C3}{285}{shift={(0.8, 0.1)}}{$V''$};
\draw[smooth, semithick] plot[tension=0.5] coordinates{(-1.2,2.5)(-.7,2.5)(-.5,2.25)(-.2,2.2)(0,2.2)(.2,2.2)(.5,2.25)(0.7,2.5)(1.2, 2.5)};

\node at (1.2, 2.3) {{\footnotesize $D$}};

\node[circle, fill=black, inner sep=0pt,minimum size=3pt] (q) at (0,2.2) {};
\node at (-.1,2.4) {{\footnotesize $q$}};


\draw[thin,densely dotted] (C1) -- (q);

\draw[name path=q_c2,thin, densely dotted] (q) -- (C2);

\draw[name path=paral,thin,opacity=0] (-1,0.42) -- (1,0.42);
\path [name intersections={of = q_c2 and paral}];
\coordinate (pprime_coord)  at (intersection-1);

\node[circle, fill=black, inner sep=0pt,minimum size=3pt] (pprime) at (pprime_coord) {};
\draw[thick,->] (q) -- (pprime);
\node at (-.4, 0.47) {{\footnotesize $p'$}};
\draw[] (-.25,0.25) rectangle(.25, .6);

\end{tikzpicture}
}%
\subfigure[] {
\begin{tikzpicture}[scale=1,on grid]
\newcommand{\basestation}[4]{%
    \node[inner sep=0pt, circle, opacity=0, #3, minimum size=2pt, fill=red] (#1) at (0*120+#2:0.2cm) {};
    \path[draw, thick, #3] 
            (0*120+#2:0.2cm) --
            (1*120+#2:0.2cm) --
            (2*120+#2:0.2cm) -- cycle;
    \node[anchor=north,inner sep=2pt] at (#1) {{\footnotesize #4}};
}   

\basestation{C1}{255}{shift={(-.8, .1)}}{$V$};
\basestation{C2}{270}{shift={(0.0, 0)}}{$V'$};
\basestation{C3}{285}{shift={(0.8, 0.1)}}{$V''$};
\draw[smooth, thin, dashed] plot[tension=0.5] coordinates{(-1.2,2.5)(-.7,2.5)(-.5,2.25)(-.2,2.2)(0,2.2)(.2,2.2)(.5,2.25)(0.7,2.5)(1.2, 2.5)};
\draw[name path=depthmap2,smooth, semithick] plot[tension=0.5] coordinates{(-1.2,2.46)(-.7,2.43)(-.6,1.87)(-.2,1.8)(.2,1.8)(.6,1.87)(.7,2.43)(1.2, 2.46)};    

\node at (1.2, 2.27) {{\footnotesize $D'$}};

\node[circle, fill=black, inner sep=0pt,minimum size=3pt] (q) at (0,2.2) {};
\node at (-.1,2.4) {{\footnotesize $q$}};
\node at (-.2,1.60) {{\footnotesize $q'$}};

\draw[name path=q_c2,thin, opacity=0] (q) -- (C2);

\path [name intersections={of = q_c2 and depthmap2}];
\coordinate (qprime_coord)  at (intersection-1);
\node[circle, fill=black, inner sep=0pt,minimum size=3pt] (qprime) at (qprime_coord) {};

\draw[name path=paral,thin,opacity=0] (-1,0.42) -- (1,0.42);
\path [name intersections={of = q_c2 and paral}];
\coordinate (pprime_coord)  at (intersection-1);
\node[circle, fill=black, inner sep=0pt,minimum size=3pt] (pprime) at (pprime_coord) {};

\draw[thick,->] (pprime) -- (qprime);
\draw[thin,densely dotted] (pprime) -- (C2);
\node at (-.4, 0.47) {{\footnotesize $p'$}};
\draw[yslant=0.25] (0.47,0.15) rectangle(.95, .5);

\draw[name path=qprime_c3,thin, opacity=0] (qprime) -- (C3);

\draw[name path=paral2,thin,opacity=0] (-1,0.44) -- (1,0.44);
\path [name intersections={of = qprime_c3 and paral2}];
\coordinate (pprime2_coord)  at (intersection-1);
\node[circle, fill=black, inner sep=0pt,minimum size=3pt] (pprime2) at (pprime2_coord) {};

\node at (.8, 0.87) {{\footnotesize $p''$}};

\draw[name path=paral3,thin,opacity=0] (-1,1.60) -- (1,1.60);
\path [name intersections={of = qprime_c3 and paral3}];
\coordinate (qprimef_coord)  at (intersection-1);
\node[circle, fill=black, inner sep=0pt,minimum size=0pt] (qprimef) at (qprimef_coord) {};

\draw[thick,->] (qprimef) -- (pprime2);
\draw[name path=qprime_c3,thin, densely dotted, ] (pprime2) -- (C3);

\draw[] (-.25,0.25) rectangle(.25, .6);

\end{tikzpicture}
}%
\subfigure[] {
\begin{tikzpicture}[scale=1,on grid]
\newcommand{\basestation}[4]{%
    \node[inner sep=0pt, circle, opacity=0, #3, minimum size=2pt, fill=red] (#1) at (0*120+#2:0.2cm) {};
    \path[draw, thick, #3] 
            (0*120+#2:0.2cm) --
            (1*120+#2:0.2cm) --
            (2*120+#2:0.2cm) -- cycle;
    \node[anchor=north,inner sep=2pt] at (#1) {{\footnotesize #4}};
}   

\basestation{C1}{255}{shift={(-.8, .1)}}{$V$};
\basestation{C2}{270}{shift={(0.0, 0)}}{$V'$};
\basestation{C3}{285}{shift={(0.8, 0.1)}}{$V''$};
\draw[smooth, thin, dashed] plot[tension=0.5] coordinates{(-1.2,2.5)(-.7,2.5)(-.5,2.25)(-.2,2.2)(0,2.2)(.2,2.2)(.5,2.25)(0.7,2.5)(1.2, 2.5)};
\draw[name path=depthmap2,smooth, thin, dashed] plot[tension=0.5] coordinates{(-1.2,2.46)(-.7,2.43)(-.6,1.87)(-.2,1.8)(.2,1.8)(.6,1.87)(.7,2.43)(1.2, 2.46)};    
\draw[name path=depthmap3,smooth, semithick] plot[tension=0.5] coordinates{(-1.2,2.43)(-.72,2.40)(-.62,1.84)(-.2,1.77)(.2,1.77)(.62,1.84)(.72,2.40)(1.2, 2.43)};    

\node at (1.2, 2.27) {{\footnotesize $D''$}};

\node[circle, fill=black, inner sep=0pt,minimum size=3pt] (q) at (0,2.2) {};
\node at (-.1,2.4) {{\footnotesize $q$}};
\node at (-.2,1.60) {{\footnotesize $q'$}};
\node at (.3,1.60) {{\footnotesize $q''$}};

\draw[name path=q_c2,thin, opacity=0] (q) -- (C2);

\path [name intersections={of = q_c2 and depthmap2}];
\coordinate (qprime_coord)  at (intersection-1);
\node[circle, fill=black, inner sep=0pt,minimum size=3pt] (qprime) at (qprime_coord) {};
\node[circle, fill=black, inner sep=0pt,minimum size=0pt,inner sep=0pt,opacity=0] (xqprime) at (qprime_coord) {};

\draw[name path=paral,thin,opacity=0] (-1,0.42) -- (1,0.42);
\path [name intersections={of = q_c2 and paral}];
\coordinate (pprime_coord)  at (intersection-1);
\node[circle, fill=black, inner sep=0pt,minimum size=3pt,opacity=0] (pprime) at (pprime_coord) {};

\draw[yslant=0.25] (0.47,0.15) rectangle(.95, .5);

\draw[name path=qprime_c3,thin, densely dotted] (xqprime) -- (C3);

\draw[name path=paral2,thin,opacity=0] (-1,0.44) -- (1,0.44);
\path [name intersections={of = qprime_c3 and paral2}];
\coordinate (pprime2_coord)  at (intersection-1);
\node[circle, fill=black, inner sep=0pt,minimum size=3pt] (pprime2) at (pprime2_coord) {};

\node at (.8, 0.87) {{\footnotesize $p''$}};

\path [name intersections={of = qprime_c3 and depthmap3}];
\coordinate (qprime2_coord)  at (intersection-1);
\node[circle, fill=black, inner sep=0pt,minimum size=3pt] (qprime2) at (qprime2_coord) {};

\draw[thick,->] (pprime2) -- (qprime2);


\end{tikzpicture}
}%

\caption {Diagram of how a 3D track is generated in three consecutive views.
a) A 3D point $q$ visible in view $V$ is projected to view $V'$ at pixel $p'$.
b) Pixel $p'$ is backprojected onto the depthmap $D'$,
creating the 3D point $q'$.
Then, the 3D point $q'$ is projected into view $V''$ at pixel $p''$.
c) Finally, pixel $p''$ is backprojected onto the depthmap $D''$, 
creating the last point in the track $q''$.
The computed track is $t=(q,q',q'')$.
Note that because the geometry remains unchanged between $V'$ and $V''$,
the points $q'$ and $q''$ are the same.
}

\label{fig:motion-track}

\end{figure}

%% file: figure_subpixel_samples.tex
\begin{figure}
\centering
\begin{tikzpicture}[scale=.65,every node/.style={minimum size=1cm},on grid]
        
     \tikzstyle{track_line}=[dotted,thick]
     \tikzstyle{pixel_center}=[circle,fill=black,inner sep=0pt,minimum size=2pt]

     \def\trackstyle{}

     \begin{scope}[
        xshift=180,every node/.append style={
        yslant=-.5,xslant=0},yslant=-.5,xslant=0
                     ]        
        \node (blue4) at (0.4,2.6) [circle,fill=blue,inner sep=0pt,minimum size=4pt,opacity=0] {};
    \end{scope}

    \begin{scope}[
        xshift=90,every node/.append style={
        yslant=-.5,xslant=0},yslant=-.5,xslant=0]        

        \node (blue3) at (1.3,2.05) [circle,fill=blue,inner sep=0pt,minimum size=4pt] {};

        \draw[track_line] (blue3) -- (blue4);

        \fill[gray!20!white,opacity=0.8] (0,0) rectangle (3,3);

        \draw[step=10mm, black] (0,0) grid (3,3);
        \draw[black,thick] (0,0) rectangle (3,3);
        \foreach \x in {.5,1.5,2.5} {
          \foreach \y in {.5,1.5,2.5} {
            \node at (\x, \y) [pixel_center] {};
          }
        }
        \node (blue3) at (blue3) [circle,fill=black,inner sep=0pt,minimum size=5pt] {};
        \node (blue3) at (blue3) [circle,fill=blue!60!lightgray,inner sep=0pt,minimum size=4pt] {};
        \node (bottom3) at (2.8, 1.) {};
    \end{scope}
    \begin{scope}[
        xshift=0,every node/.append style={
        yslant=-.5,xslant=0},yslant=-.5,xslant=0]

        \node (blue2) at (1.8,1.4) [circle,fill=blue,inner sep=0pt,minimum size=4pt] {};

        \draw[track_line] (blue2) -- (blue3);
        \fill[gray!20!white,opacity=0.8] (0,0) rectangle (3,3);

        \draw[step=10mm, black] (0,0) grid (3,3);
        \draw[black,thick] (0,0) rectangle (3,3);
        \foreach \x in {.5,1.5,2.5} {
          \foreach \y in {.5,1.5,2.5} {
            \node at (\x, \y) [pixel_center] {};
          }
        }
        \node (blue2) at (blue2) [circle,fill=black,inner sep=0pt,minimum size=5pt] {};

        \node (blue2) at (blue2) [circle,fill=blue!60!lightgray,inner sep=0pt,minimum size=4pt] {};

        \node (bottom2) at (2.8, 1.) {};

    \end{scope}

    \begin{scope}[
        xshift=-90,every node/.append style={
        yslant=-.5,xslant=0},yslant=-.5,xslant=0]

        \node (red1) at (2.5,0.7) [circle,fill=red,inner sep=0pt,minimum size=4pt] {};

        \draw[track_line] (red1) -- (blue2);
        \fill[gray!20!white,opacity=0.8] (0,0) rectangle (3,3);

        \draw[step=10mm, black] (0,0) grid (3,3);
        \draw[black, thick] (0,0) rectangle (3,3);
        \foreach \x in {.5,1.5,2.5} {
          \foreach \y in {.5,1.5,2.5} {
            \node at (\x, \y) [pixel_center] {};
          }
        }
        \node (red1) at (red1) [circle,fill=black,inner sep=0pt,minimum size=5pt] {};
        \node (red1) at (red1) [circle,fill=green!80!white,inner sep=0pt,minimum size=4pt] {};

        \node (bottom1) at (2.8, 1.) {};

    \end{scope}




     \begin{scope}[
        xshift=-180,every node/.append style={
        yslant=-.5,xslant=0},yslant=-.5,xslant=0
                     ]        
        \node (red0) at (2.6,0.2) [circle,fill=red,inner sep=0pt,minimum size=4pt,opacity=0] {};
        \draw[track_line] (red0) -- (red1);

    \end{scope}

    \node[below of=bottom3] {\small $V_{j+1}$};
    \node[below of=bottom2] {\small $V_{j}$};
    \node[below of=bottom1] {\small $V_{j-1}$};

\end{tikzpicture}

\caption{
Projected temporal color profiles of a 3D track $t$ into three views.
The views are represented by a pixel grid, with the pixel centers marked as
black dots.
The projected temporal color profiles are defined by a real-valued projected position $p^t_j$
into view $j$ and a time-varying, regularized color $y^t_j$.
The projected profile is shown as a sequence of colored circles, projected 
on each view, linked by a dashed line.
}
\label{fig:subpixel-samples}
\end{figure}

%% file: figure_subpixel_recon.tex

\begin{figure}
\centering
\subfigure[Projected color profiles]{
\begin{tikzpicture}[scale=0.7,every node/.style={minimum size=1cm},on grid]
\tikzstyle{pixel_center}=[circle,fill=black,inner sep=0pt,minimum size=2pt]

\begin{scope}
    \draw[white] (0,1) grid (5,4);

    \def\op{0.8}

    \draw[step=10mm, black, thin, dashed,black!50!gray] (0.5,.5) grid (4.5,4.5);
    \draw[black,thick, step=10mm,black!50!gray] (1,1) grid (4,4);

    \node at (1.5, 1.5) [pixel_center] (c11) {};
    \node at (2.5, 1.5) [pixel_center] (c21) {};
    \node at (2.5, 2.5) [pixel_center] (c22) {};
    \node at (1.5, 2.5) [pixel_center] (c12) {};

    \node at (1.5, 3.5) [pixel_center] (c13) {};
    \node at (2.5, 3.5) [pixel_center] (c23) {};
    \node at (3.5, 3.5) [pixel_center] (c33) {};
    \node at (3.5, 2.5) [pixel_center] (c32) {};
    \node at (3.5, 1.5) [pixel_center] (c31) {};

    \def\opx{1}

     \tikzstyle{arrow_style}=[black, semithick,->,opacity=1]

    \node[circle,fill=black,inner sep=1pt,minimum size=5pt,opacity=\opx] (S1) at (1.80, 2.15) {};
    \node[circle,fill=blue,inner sep=1pt,minimum size=4pt,opacity=\opx] (S1) at (1.80, 2.15) {};
    \draw[arrow_style] (S1) -- (c11);
    \draw[arrow_style] (S1) -- (c21);
    \draw[arrow_style] (S1) -- (c12);
    \draw[arrow_style] (S1) -- (c22);

    \node[circle,fill=black,inner sep=1pt,minimum size=5pt,opacity=\opx] (S2) at (3.3, 3.1) {};
    \node[circle,fill=blue!80!green,inner sep=1pt,minimum size=4pt,opacity=\opx] (S2) at (3.3, 3.1) {};
    \draw[arrow_style] (S2) -- (c22);
    \draw[arrow_style] (S2) -- (c23);
    \draw[arrow_style] (S2) -- (c32);
    \draw[arrow_style] (S2) -- (c33);

    \node[circle,fill=black,inner sep=1pt,minimum size=5pt,opacity=\opx] (S3) at (1.9, 3.2) {};
    \node[circle,fill=blue!90!pink,inner sep=1pt,minimum size=4pt,opacity=\opx] (S3) at (1.9, 3.2) {};
    \draw[arrow_style] (S3) -- (c12);
    \draw[arrow_style] (S3) -- (c13);
    \draw[arrow_style] (S3) -- (c23);
    \draw[arrow_style] (S3) -- (c22);

    \node[circle,fill=black,inner sep=1pt,minimum size=5pt,opacity=\opx] (S4) at (3.2,1.9) {};
    \node[circle,fill=green!80!white,inner sep=1pt,minimum size=4pt,opacity=\opx] (S4) at (3.2,1.9) {};
    \draw[arrow_style] (S4) -- (c21);
    \draw[arrow_style] (S4) -- (c22);
    \draw[arrow_style] (S4) -- (c31);
    \draw[arrow_style] (S4) -- (c32);

\end{scope}
\end{tikzpicture}}
\hspace{5pt}
\subfigure[Reconstructed image]{

\begin{tikzpicture}[scale=0.7,every node/.style={minimum size=1cm},on grid]
\tikzstyle{pixel_center}=[circle,fill=black,inner sep=0pt,minimum size=2pt]
\draw[white] (0,1) grid (5,4);

\begin{scope}
    \node at (1.5, 1.5) [pixel_center] (c11) {};
    \node at (2.5, 1.5) [pixel_center] (c21) {};
    \node at (2.5, 2.5) [pixel_center] (c22) {};
    \node at (1.5, 2.5) [pixel_center] (c12) {};

    \node at (1.5, 3.5) [pixel_center] (c13) {};
    \node at (2.5, 3.5) [pixel_center] (c23) {};
    \node at (3.5, 3.5) [pixel_center] (c33) {};
    \node at (3.5, 2.5) [pixel_center] (c32) {};
    \node at (3.5, 1.5) [pixel_center] (c31) {};

    \def\op{0.4}
    \draw[fill=blue!65!green,opacity=\op] (1, 1) rectangle(2, 2);
    \draw[fill=blue!86!green,opacity=\op] (1, 2) rectangle(2, 3);
    \draw[fill=blue!53!green,opacity=\op] (2, 1) rectangle(3, 2);
    \draw[fill=blue!75!green,opacity=\op] (2, 2) rectangle(3, 3);
    \draw[fill=blue!20!green,opacity=\op] (3, 1) rectangle(4, 2);
    \draw[fill=blue!40!green,opacity=\op] (3, 2) rectangle(4, 3);
    \draw[fill=blue!80!green,opacity=\op] (3, 3) rectangle(4, 4);
    \draw[fill=blue!90!green,opacity=\op] (2, 3) rectangle(3, 4);
    \draw[fill=blue,opacity=\op] (1, 3) rectangle(2, 4);

    \draw[step=10mm, black!50!gray, thin, dashed] (0.5,.5) grid (4.5,4.5);
    \draw[black!50!gray,thick, step=10mm] (1,1) grid (4,4);

\end{scope}

\end{tikzpicture}
}

\caption{Visualization of the output frame reconstruction algorithm from projected
color profiles.
Left: Projected color profiles at a given view shown as colored dots
in the output frame
with their bilinear interpolation weights shown as arrows from the projected sample to pixel centers.
Right: We reconstruct an image that minimizes the bilinear interpolation
residuals of the projected color profiles.
}
\label{fig:subpixel-recon}

\end{figure} 

%% file: implementation.tex
\section{Implementation}
\label{sec:implementation}
The photo collections used in our system consist of publicly available Picasa and Panoramio images.
For a single landmark, the 3D reconstructions contain up to 25K photos,
and the input sequences filtered with the camera
selection criteria of~\cite{TimelapseMining} contain
between 500 and 2200 photos.
We generate virtual camera paths containing between 100 and 200
frames.

The weights for the depthmap computation are $\alpha=0.4$ and the 
temporal binary weight is defined as $\beta_{j,j'}= k_1 \max{(1-|j'-j|/k_2, 0)}$ with $k_1 = 30$ and $k_2 = 8$.
The scale parameter of the Huber loss used for $E^s$ and $E^t$ is $0.1$ disparity values.
For appearance regularization, we use the Huber loss for $\delta_d$ and $\delta_t$ with scale parameter of $1^{-4}$, \ie, about $1/4$ of a pixel value.
Finally, the temporal regularization weight is $\lambda=25$.
We use Ceres Solver~\cite{ceres-solver} to solve for the optimized color profiles,
that we solve per color channel independently.

Our multi-threaded CPU implementation runs on a single workstation with 12 cores
and 48Gb memory in 4 hours and 10 minutes for a 100 frame sequence.
The breakdown is the following: 151 minutes for depthmap initialization, 30 minutes for joint depthmap optimization,
55 minutes for 3D track generation and regularization, and 25 minutes for video reconstruction.
We compute the output sequences at an image resolution of $800\times 600$, with a depthmap resolution of $400 \times 300$.
Our execution time is dominated by the cost volume computation for all the views,
and we subsample the support sets $\mathcal{S}_j$ to contain at most $100$ images without noticeable detrimental effects.

%% file: results.tex
\section{Results}
\label{sec:results}
\setlength{\textfloatsep}{10pt}

\begin{figure}[t]
\begin{tikzpicture}
    \node[anchor=south west,inner sep=0] (image) at (0,0) {\includegraphics[width=0.49\linewidth]{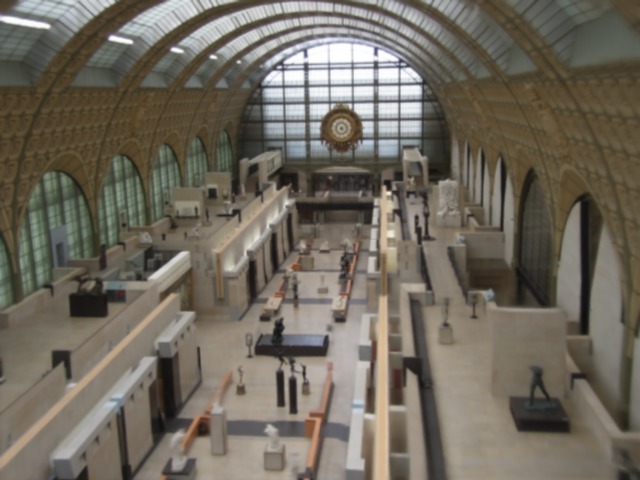}};
    \begin{scope}[x={(image.south east)},y={(image.north west)}]
        \node[anchor=south west,inner sep=0] (image) at (0.49,0.49) {\includegraphics[width=0.245\linewidth]{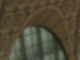}};
        \draw [semithick](0.031250,1-0.302083) rectangle (0.156250,1-0.427083);
        \draw [semithick](0.49,0.49) rectangle (0.99,0.99);
        \draw[semithick,dashed] (0.031250,1-0.302083) -- (0.49,0.99);
        \draw[semithick,dashed] (0.031250,1-0.427083) -- (0.49,0.49);
    \end{scope}
\end{tikzpicture}
\begin{tikzpicture}
    \node[anchor=south west,inner sep=0] (image) at (0,0) {\includegraphics[width=0.49\linewidth]{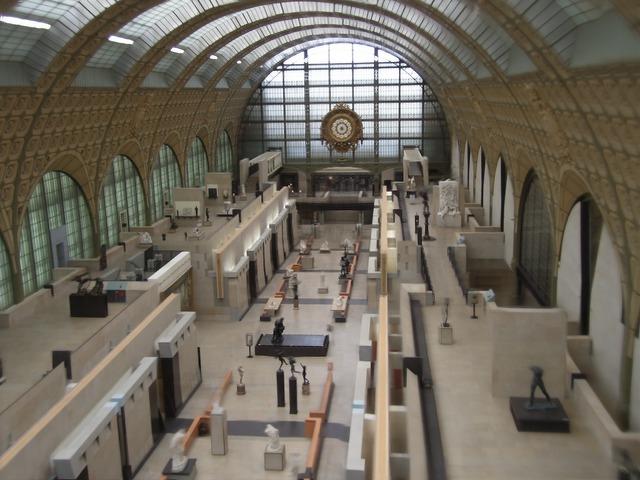}};
    \begin{scope}[x={(image.south east)},y={(image.north west)}]
        \node[anchor=south west,inner sep=0] (image) at (0.49,0.49) {\includegraphics[width=0.245\linewidth]{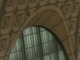}};
        \draw [semithick](0.031250,1-0.302083) rectangle (0.156250,1-0.427083);
        \draw [semithick](0.49,0.49) rectangle (0.99,0.99);
        \draw[semithick,dashed] (0.031250,1-0.302083) -- (0.49,0.99);
        \draw[semithick,dashed] (0.031250,1-0.427083) -- (0.49,0.49);
    \end{scope}
\end{tikzpicture}

\caption{
Comparison of two methods for output frame reconstruction from projected profiles for the Mus\'{e}e D'Orsay scene.
Left: baseline method based on Gaussian kernel splatting, with kernel radius $\sigma=1$.
Right:  our reconstruction approach.
The baseline method produces a blurred reconstruction,
whereas the proposed approach recovers high frequency details in the output frame.
}
\label{fig:subpixel-baseline}
\end{figure}
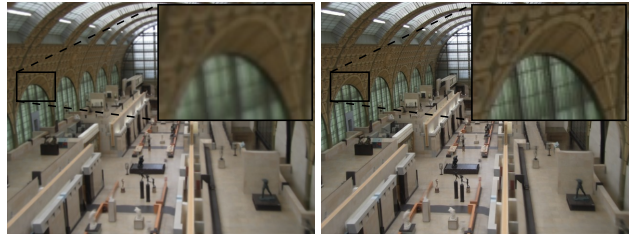

\begin{figure*}[ht!]
\hspace{.9pt}
\begin{tikzpicture}
    \node[anchor=south west,inner sep=0] (image) at (0,0) {\includegraphics[width=0.34\linewidth]{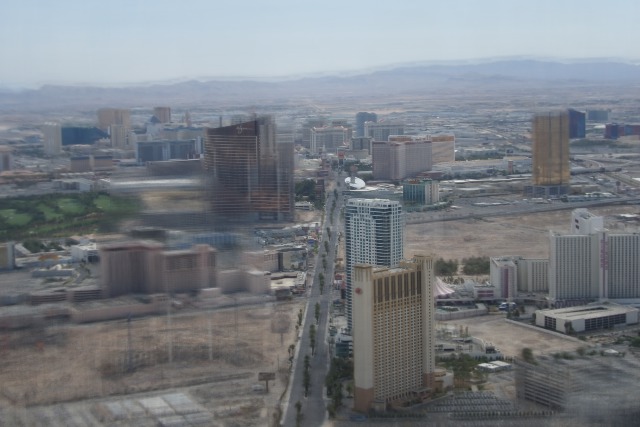}};
    \begin{scope}[x={(image.south east)},y={(image.north west)}]
        \node[anchor=south west,inner sep=0] (image) at (1.01,0.00) {\includegraphics[width=0.1515\linewidth]{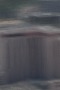}};
        \draw [semithick](0.159375,1-0.501171) rectangle (0.253125,1-0.711944);
        \draw [thick](1.01,0.0) rectangle (1.458,1.0);
        \draw[semithick,dashed] (0.159375,1-0.501171) -- (1.01,1);
        \draw[semithick,dashed] (0.159375,1-0.711944) -- (1.01,0);
    \end{scope}
\end{tikzpicture}
\begin{tikzpicture}
    \node[anchor=south west,inner sep=0] (image) at (0,0) {\includegraphics[width=0.34\linewidth]{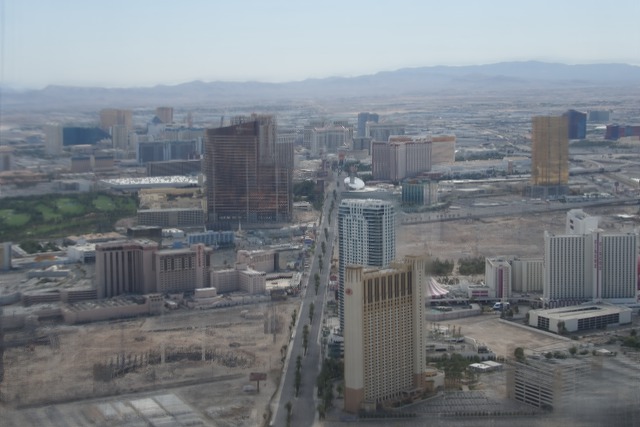}};
    \begin{scope}[x={(image.south east)},y={(image.north west)}]
        \node[anchor=south west,inner sep=0] (image) at (1.01,0.00) {\includegraphics[width=0.1515\linewidth]{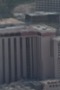}};
        \draw [semithick](0.159375,1-0.501171) rectangle (0.253125,1-0.711944);
        \draw [thick](1.01,0.0) rectangle (1.458,1.0);
        \draw[semithick,dashed] (0.159375,1-0.501171) -- (1.01,1);
        \draw[semithick,dashed] (0.159375,1-0.711944) -- (1.01,0);
    \end{scope}
\end{tikzpicture}%
\vspace{-4pt}
\\
\subfigure[Static depthmap]{
\begin{tikzpicture}
    \node[anchor=south west,inner sep=0] (image) at (0,0) {\includegraphics[width=0.34\linewidth]{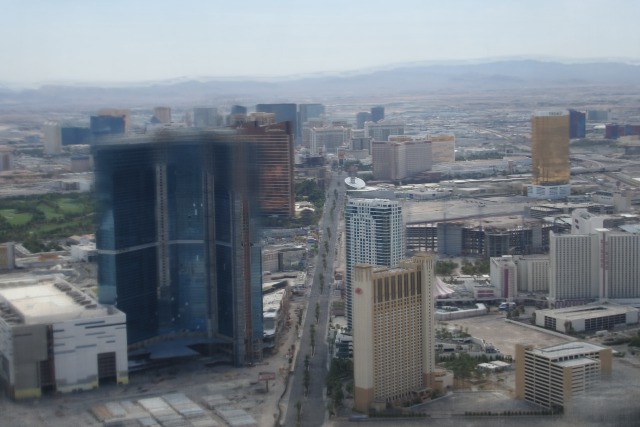}};
    \begin{scope}[x={(image.south east)},y={(image.north west)}]
        \node[anchor=south west,inner sep=0] (image) at (1.01,0.00) {\includegraphics[width=0.1515\linewidth]{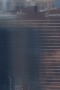}};
        \draw [semithick](0.343750,1-0.2693151) rectangle (0.437500,1-0.480094);
        \draw [thick](1.01,0.0) rectangle (1.458,1.0);
        \draw[semithick,dashed] (0.343750,1-0.2693151) -- (1.01,1);
        \draw[semithick,dashed] (0.343750,1-0.480094) -- (1.01,0);
    \end{scope}
\end{tikzpicture}}%
\subfigure[Time-varying depthmap]{
\begin{tikzpicture}
    \node[anchor=south west,inner sep=0] (image) at (0,0) {\includegraphics[width=0.34\linewidth]{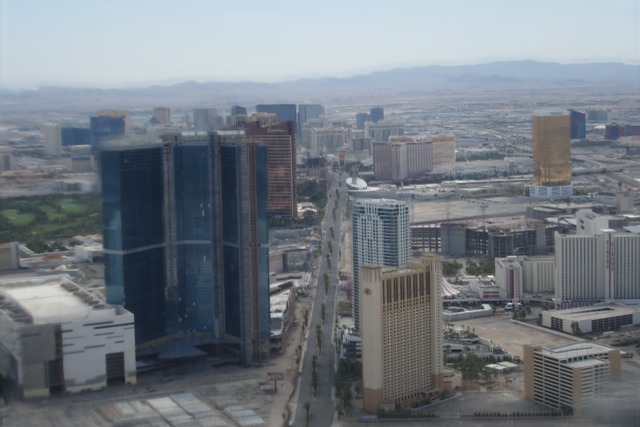}};
    \begin{scope}[x={(image.south east)},y={(image.north west)}]
        \node[anchor=south west,inner sep=0] (image) at (1.01,0.00) {\includegraphics[width=0.1515\linewidth]{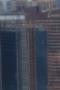}};
        \draw [semithick](0.343750,1-0.2693151) rectangle (0.437500,1-0.480094);
        \draw [thick](1.01,0.0) rectangle (1.458,1.0);
        \draw[semithick,dashed] (0.343750,1-0.2693151) -- (1.01,1);
        \draw[semithick,dashed] (0.343750,1-0.480094) -- (1.01,0);
    \end{scope}
\end{tikzpicture}}%
\caption{
Comparison of output time-lapse frames for two different timestamps for the Las Vegas sequence.
a) Using a static depthmap solved with a discrete MRF as in~\cite{TimelapseMining}.
b) Using our time-varying, temporally consistent depthmaps.
The static depthmap is not able to stabilize the input images for the whole time-lapse,
creating blurry artifacts where the geometry changes significantly.
Thanks to the time-varying depthmap, our 3D time-lapses are sharp
over the whole sequence.
}
\label{fig:static-depthmap-comp}
\end{figure*}
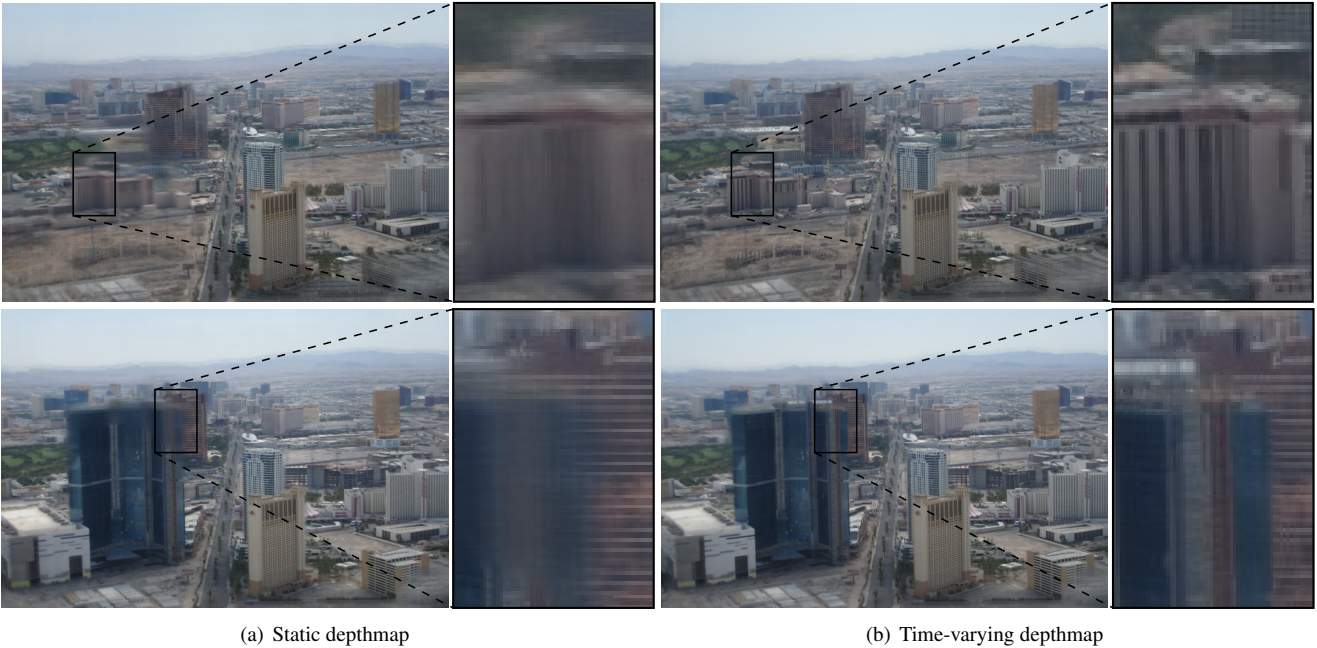

\begin{figure*}
\centering
\vspace{-10pt}
\subfigure[Flatiron Building, New York]{%
\includegraphics[width=0.315\linewidth]{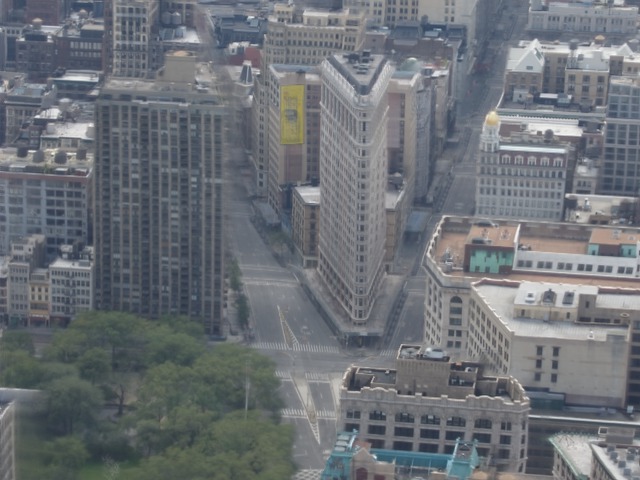}
\includegraphics[width=0.315\linewidth]{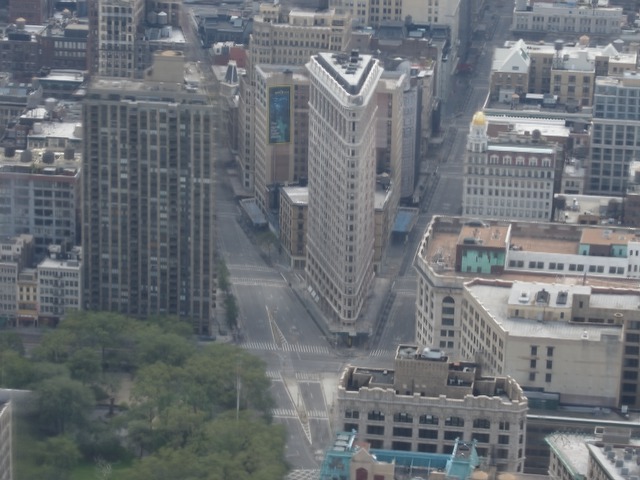}
\includegraphics[width=0.315\linewidth]{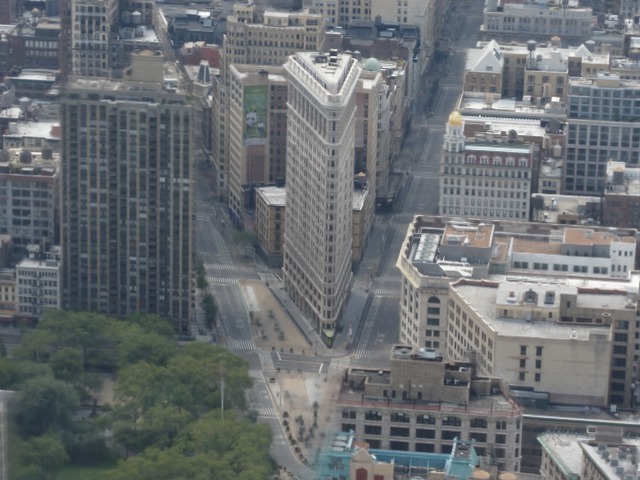}
}\\
\vspace{-6pt}
\subfigure[Lombard Street, San Francisco]{%
\includegraphics[width=0.315\linewidth]{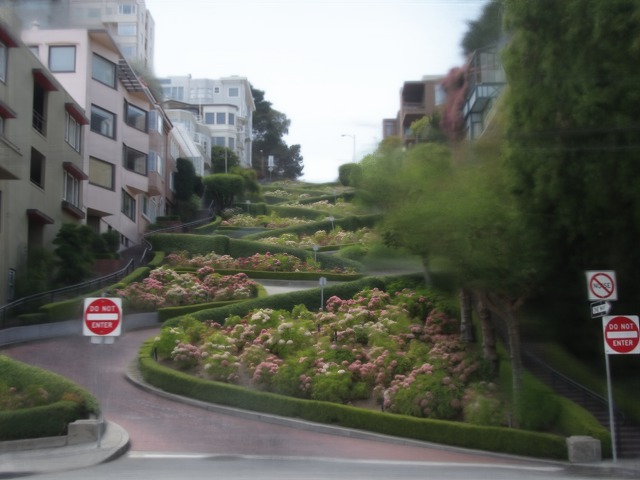}
\includegraphics[width=0.315\linewidth]{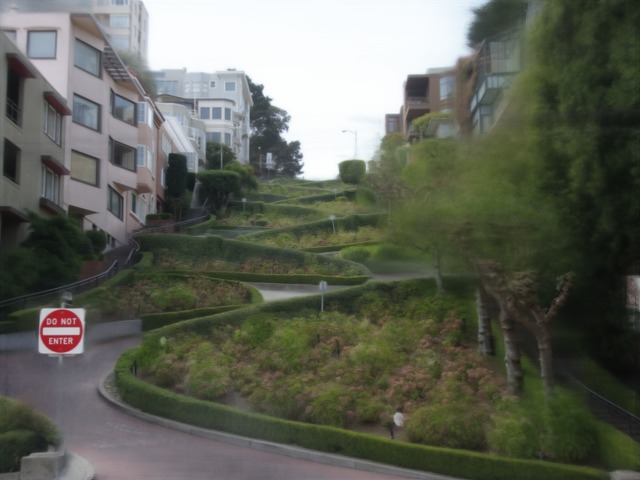}
\includegraphics[width=0.315\linewidth]{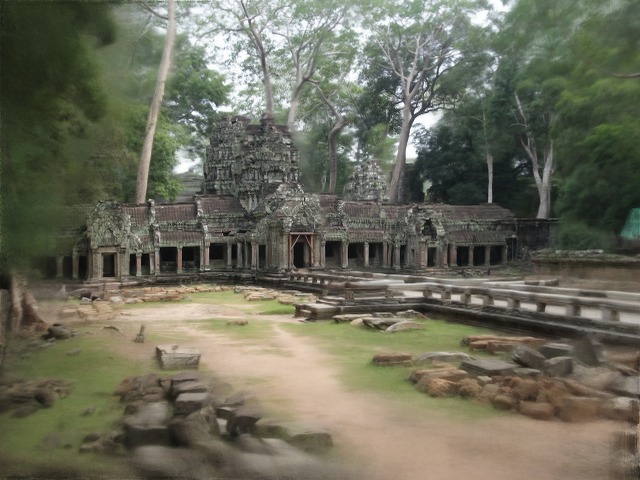}
}\\
\vspace{-6pt}
\subfigure[Ta Prohm, Cambodia]{%
\includegraphics[width=0.315\linewidth]{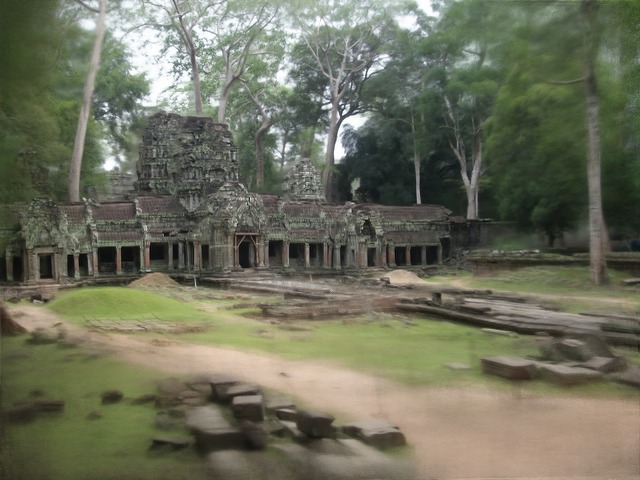}
\includegraphics[width=0.315\linewidth]{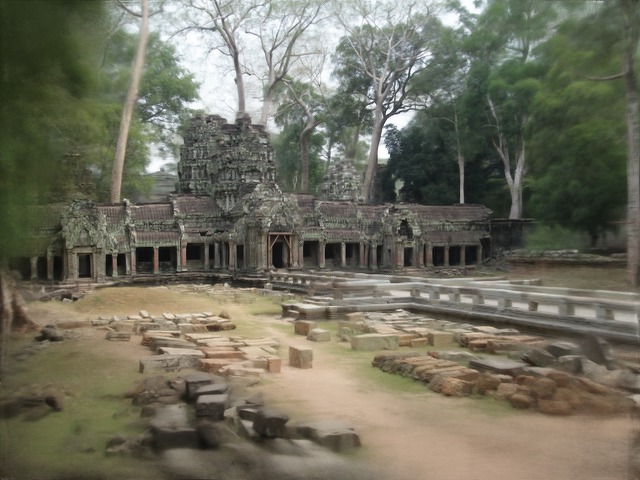}
\includegraphics[width=0.315\linewidth]{subpixel_recon_0199.jpg}
}\\
\vspace{-6pt}
\subfigure[Palette Springs, Yellowstone]{%
\includegraphics[width=0.315\linewidth]{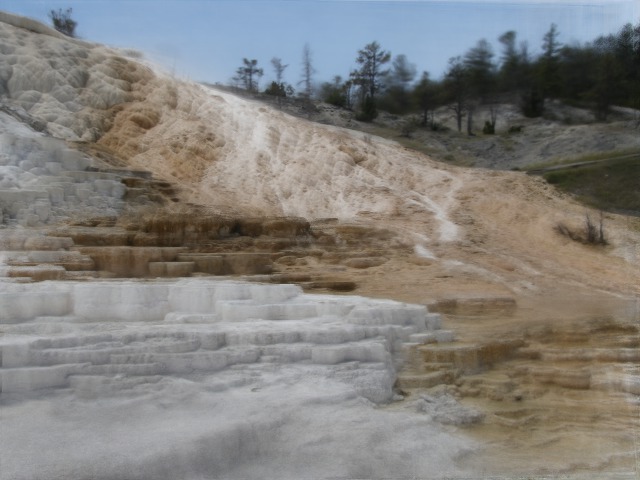}
\includegraphics[width=0.315\linewidth]{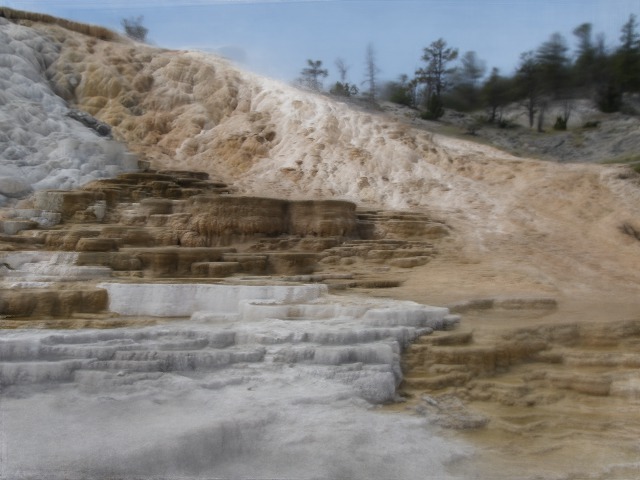}
\includegraphics[width=0.315\linewidth]{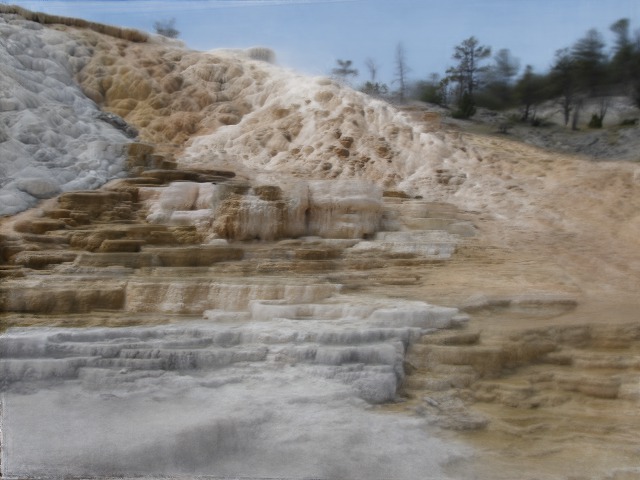}
}
  \caption{
Frames from example 3D time-lapses, with time spans of several years and subtle camera motions.
Sequences a), c) and d) contain an orbit camera path, while b) contains a camera ``push''.
Parallax effects are best seen in the video available at the project website~\cite{3DTimelapseWebsite}.
Limitations of our system include blurry artifacts in the foreground, like in c) and d).
}
\label{fig:results}
\end{figure*}

We generated high-quality 3D time-lapse videos for 14 scenes, spanning time periods between 4 and 10 years.
Figure~\ref{fig:results} shows sample frames from four different scenes.
We refer the reader to the supplementary video~\cite{3DTimelapseWebsite} to better appreciate
the changes in the scenes and the parallax effects in our
3D time-lapses.

Figure~\ref{fig:static-depthmap-comp} shows a comparison of our 3D time-lapse
for the Las Vegas sequence with the result of previous work~\cite{TimelapseMining},
that was noted as a failure case due to changing scene geometry. 
Our 3D time-lapse result eliminates the blurry artifacts,
as the time-varying depthmap recovers the building construction process accurately.

We also compare our output frame reconstruction approach with a baseline method
that uses splatting of the projected color profiles with Gaussian weights.
Each projected profile sample contributes its color to nearby pixels 
with a weight based on the distance to the pixel center. 
Figure~\ref{fig:subpixel-baseline} shows that the baseline produces
blurred results whereas  our approach recovers
high frequency details in the output frame.

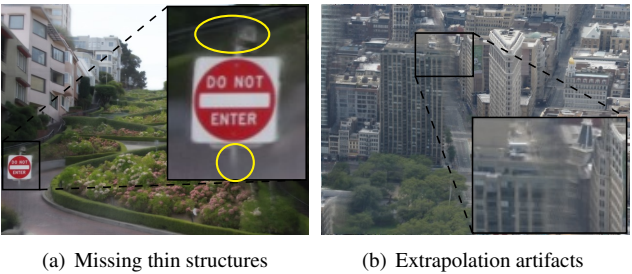
\begin{figure}
\subfigure[Missing thin structures]{
\label{fig:thin-structure}
\begin{tikzpicture}
    \node[anchor=south west,inner sep=0] (image) at (0,0) {\includegraphics[width=0.49\linewidth]{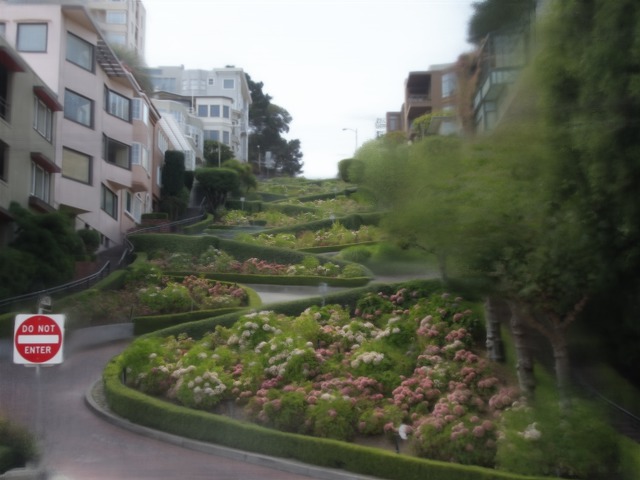}};
    \begin{scope}[x={(image.south east)},y={(image.north west)}]
        \node[anchor=south west,inner sep=0] (image) at (0.54,0.24) {\includegraphics[height=0.275625\linewidth]{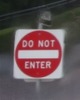}};
        \draw [semithick] (0.005,1-0.593750) rectangle (0.125000,1-0.802083);
        \draw [semithick] (0.54,0.24) rectangle (0.99,0.99);
        \draw [semithick,dashed] (0.005,1-0.593750) -- (0.54,0.99);
        \draw [semithick,dashed] (0.125000,1-0.802083) -- (0.99,0.24);
      \draw [yellow,thick] (0.75,0.87) ellipse (0.12 and 0.08);
      \draw [yellow,thick] (0.76,0.315) ellipse (0.06 and 0.08);

    \end{scope}
\end{tikzpicture}
}%
\subfigure[Extrapolation artifacts	]{
\label{fig:extrapolation}
\begin{tikzpicture}
    \node[anchor=south west,inner sep=0] (image) at (0,0) {\includegraphics[width=0.49\linewidth]{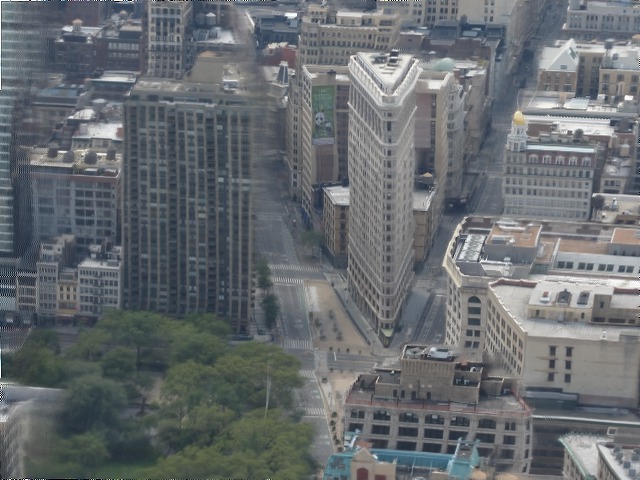}};
    \begin{scope}[x={(image.south east)},y={(image.north west)}]
        \node[anchor=south west,inner sep=0] (image) at (0.49,0.01) {\includegraphics[width=0.245\linewidth]{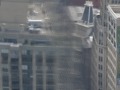}};
        \draw [semithick] (0.304688,1-0.125000) rectangle (0.492188,1-0.312500);
        \draw [semithick] (0.49,0.01) rectangle (0.99,0.51);
        \draw [semithick,dashed] (0.492188,1-0.125000) -- (0.99,0.51);
        \draw [semithick,dashed] (0.304688,1-0.312500) -- (0.49,0.01);
    \end{scope}
\end{tikzpicture}
}

\caption{
Examples of failure cases in our system. 
a) The street sign is not fully reconstructed in the Lombard Street sequence.
b) An extended camera orbit contains a virtual camera far from the set of input cameras causing blurry artifacts in the Flatiron Building dataset. 
}
\label{fig:failure-case}
\end{figure}

\setlength{\textfloatsep}{10pt}

\subsection{Limitations}
We observed a few failure cases in our system.
Inaccurate depthmaps create blurring or shearing artifacts, especially if close objects are present.
For example, in the Lombard Street sequence shown in Figure~\ref{fig:thin-structure},
the system fails to reconstruct thin structures, blurring them away.
Recovering more accurate, time-varying geometry from Internet photo collections
is an area of future work.

Our system also generates artifacts when extrapolating the input photo collection.
This happens when a camera looks at a surface not visible in any input photo.
For example, in Figure~\ref{fig:extrapolation} a view is synthesized for a camera
outside the convex hull of reconstructed cameras, showing a face of a building that is
not visible from any photo.
Future work could consider using visibility information to constrain the virtual camera paths like in~\cite{ParallaxPhotography}.

Our technique is limited to reconstructing 3D time-lapses given pre-specified camera paths.
Future work includes enabling interactive visualizations of these photorealistic 3D time-lapses.

%% file: conclusion.tex
\section{Conclusion}

In this paper we introduce a method to reconstruct 3D time-lapse videos
from Internet photos where a virtual camera moves continuously in time and space.
Our method involves solving for time-varying depthmaps,
regularizing 3D point color profiles over time,
and reconstructing high quality, hole-free output frames.
By using cinematographic camera paths, we generate time-lapse videos
with compelling parallax effects.

\section*{Acknowledgements}
The research was supported in part by the National Science Foundation (IIS-1250793), the Animation Research Labs, and Google.

\clearpage

%% file: submission.bbl
\begin{thebibliography}{}\itemsep=-1pt

\bibitem{agarwal2011romeinaday}
S.~Agarwal, Y.~Furukawa, N.~Snavely, I.~Simon, B.~Curless, S.~M. Seitz, and
  R.~Szeliski.
\newblock Building rome in a day.
\newblock {\em Communications of the ACM}, 54(10):105--112, 2011.

\bibitem{ceres-solver}
S.~Agarwal, K.~Mierle, and Others.
\newblock Ceres {S}olver.
\newblock \url{http://ceres-solver.org}.

\bibitem{ComputationalTimelapse}
E.~P. Bennett and L.~McMillan.
\newblock Computational time-lapse video.
\newblock In {\em ACM SIGGRAPH 2007 Papers}, SIGGRAPH '07, New York, NY, USA,
  2007. ACM.

\bibitem{ExtremeIceSurvey}
{\relax Earth Vision Institute}.
\newblock Extreme {I}ce {S}urvey.
\newblock {\small \url{http://extremeicesurvey.org/}}, 2007.

\bibitem{hauagge_bmvc2014_outdoor}
D.~Hauagge, S.~Wehrwein, P.~Upchurch, K.~Bala, and N.~Snavely.
\newblock Reasoning about photo collections using models of outdoor
  illumination.
\newblock In {\em Proceedings of BMVC}, 2014.

\bibitem{Kang04extractingviewdependent}
S.~B. Kang and R.~Szeliski.
\newblock Extracting view-dependent depth maps from a collection of images.
\newblock {\em International Journal of Computer Vision}, 58:139--163, 2004.

\bibitem{Photobios}
I.~Kemelmacher-Shlizerman, E.~Shechtman, R.~Garg, and S.~M. Seitz.
\newblock Exploring photobios.
\newblock In {\em ACM SIGGRAPH 2011 Papers}, SIGGRAPH '11, pages 61:1--61:10,
  New York, NY, USA, 2011. ACM.

\bibitem{FirstPersonHyperlapse}
J.~Kopf, M.~F. Cohen, and R.~Szeliski.
\newblock First-person hyper-lapse videos.
\newblock {\em ACM Trans. Graph.}, 33(4):78:1--78:10, July 2014.

\bibitem{CoherentIntrinsicImages}
P.-Y. Laffont, A.~Bousseau, S.~Paris, F.~Durand, and G.~Drettakis.
\newblock Coherent intrinsic images from photo collections.
\newblock {\em ACM Transactions on Graphics (SIGGRAPH Asia Conference
  Proceedings)}, 31, 2012.

\bibitem{LaforetTimelapseTutorial}
V.~Laforet.
\newblock Time {L}apse {I}ntro: {P}art {I}.
\newblock {\small
  \url{http://blog.vincentlaforet.com/2013/04/27/time-lapse-intro-part-i/}}.

\bibitem{Larsen2007}
E.~Larsen, P.~Mordohai, M.~Pollefeys, and H.~Fuchs.
\newblock Temporally consistent reconstruction from multiple video streams
  using enhanced belief propagation.
\newblock In {\em Computer Vision, 2007. ICCV 2007. IEEE 11th International
  Conference on}, pages 1--8, Oct 2007.

\bibitem{TimelapseMining}
R.~Martin-Brualla, D.~Gallup, and S.~M. Seitz.
\newblock Time-lapse mining from internet photos.
\newblock {\em ACM Trans. Graph.}, 34(4):62:1--62:8, July 2015.

\bibitem{MatzenECCV14}
K.~Matzen and N.~Snavely.
\newblock Scene chronology.
\newblock In {\em Proc. European Conf. on Computer Vision}, 2014.

\bibitem{DTAM}
R.~A. Newcombe, S.~Lovegrove, and A.~Davison.
\newblock Dtam: Dense tracking and mapping in real-time.
\newblock In {\em Computer Vision (ICCV), 2011 IEEE International Conference
  on}, pages 2320--2327, Nov 2011.

\bibitem{3DTimelapseWebsite}
{\relax Project Website}.
\newblock \url{http://grail.cs.washington.edu/projects/timelapse3d}.

\bibitem{MotionDenoising}
M.~Rubinstein, C.~Liu, P.~Sand, F.~Durand, and W.~T. Freeman.
\newblock Motion denoising with application to time-lapse photography.
\newblock In {\em Proceedings of the 2011 IEEE Conference on Computer Vision
  and Pattern Recognition}, CVPR '11, pages 313--320, Washington, DC, USA,
  2011. IEEE Computer Society.

\bibitem{SchindlerCVPR2010}
G.~Schindler and F.~Dellaert.
\newblock Probabilistic temporal inference on reconstructed 3d scenes.
\newblock In {\em Computer Vision and Pattern Recognition (CVPR), 2010 IEEE
  Conference on}, pages 1410--1417. IEEE, 2010.

\bibitem{SchindlerCVPR2007}
G.~Schindler, F.~Dellaert, and S.~B. Kang.
\newblock Inferring temporal order of images from 3d structure.
\newblock In {\em Computer Vision and Pattern Recognition, 2007. CVPR '07. IEEE
  Conference on}, pages 1--7, June 2007.

\bibitem{MultiViewStereoSeitz06}
S.~Seitz, B.~Curless, J.~Diebel, D.~Scharstein, and R.~Szeliski.
\newblock A comparison and evaluation of multi-view stereo reconstruction
  algorithms.
\newblock In {\em Computer Vision and Pattern Recognition, 2006 IEEE Computer
  Society Conference on}, volume~1, pages 519--528, June 2006.

\bibitem{VisualTuringTest}
Q.~Shan, R.~Adams, B.~Curless, Y.~Furukawa, and S.~Seitz.
\newblock The visual turing test for scene reconstruction.
\newblock In {\em 3D Vision - 3DV 2013, 2013 International Conference on},
  pages 25--32, June 2013.

\bibitem{SceneSummarization}
I.~Simon, N.~Snavely, and S.~Seitz.
\newblock Scene summarization for online image collections.
\newblock In {\em Computer Vision, 2007. ICCV 2007. IEEE 11th International
  Conference on}, pages 1--8, Oct 2007.

\bibitem{FindingPaths}
N.~Snavely, R.~Garg, S.~M. Seitz, and R.~Szeliski.
\newblock Finding paths through the world's photos.
\newblock {\em ACM Transactions on Graphics (Proceedings of SIGGRAPH 2008)},
  27(3):11--21, 2008.

\bibitem{VideoDepthmapRecovery2009}
G.~Zhang, J.~Jia, T.-T. Wong, and H.~Bao.
\newblock Consistent depth maps recovery from a video sequence.
\newblock {\em Pattern Analysis and Machine Intelligence, IEEE Transactions
  on}, 31(6):974--988, June 2009.

\bibitem{ZhangSpaceTimeStereo}
L.~Zhang, B.~Curless, and S.~Seitz.
\newblock Spacetime stereo: shape recovery for dynamic scenes.
\newblock In {\em Computer Vision and Pattern Recognition, 2003. Proceedings.
  2003 IEEE Computer Society Conference on}, volume~2, pages II--367--74 vol.2,
  June 2003.

\bibitem{ParallaxPhotography}
K.~C. Zheng, A.~Colburn, A.~Agarwala, M.~Agrawala, D.~Salesin, B.~Curless, and
  M.~F. Cohen.
\newblock Parallax photography: Creating 3d cinematic effects from stills.
\newblock In {\em Proceedings of Graphics Interface 2009}, GI '09, pages
  111--118, Toronto, Ont., Canada, Canada, 2009. Canadian Information
  Processing Society.

\end{thebibliography}
